\documentclass[a4paper,fleqn]{cas-sc}
\usepackage{lineno}

\usepackage[authoryear,sort&compress]{natbib}
\usepackage{float}
\usepackage{tabularx}
\usepackage{threeparttable}
\usepackage[expansion=false]{microtype}
\usepackage{xurl}
\usepackage{tikz}
\usepackage{enumitem}
\usepackage{array}
\usepackage{multirow}
\renewcommand{\arraystretch}{1.35}
\newcommand{\panellabelfont}{\footnotesize\bfseries}

\newsavebox{\panelbox}
\newlength{\panelht}
\newcommand{\abovepanel}[3]{%
  \sbox{\panelbox}{\includegraphics[height=#1]{#2}}%
  \begin{minipage}[t]{\wd\panelbox}
    \raggedright{\panellabelfont #3}\par\vspace{2pt}%
    \usebox{\panelbox}%
  \end{minipage}%
}

\begin{document}

\shorttitle{National-scale tree species composition mapping in Switzerland}
\ExplSyntaxOn
\bool_gset_true:N \g_stm_nologo_bool
\ExplSyntaxOff

\shortauthors{Zhou et al.}

\title[mode=title]{National-scale mapping of tree species composition in Switzerland from high-resolution aerial imagery and multimodal data}

\author[1]{Yixin Zhou}[orcid=0009-0005-6619-0411]
\cormark[1]
\ead{yixizhou@student.ethz.ch}
\author[2,4]{Luca Ferrari}
\author[2,3]{Verena C. Griess}
\author[5]{Nataliia Rehush}
\author[1]{Konrad Schindler}
\author[2,4]{Mirela Beloiu}

\affiliation[1]{organization={Photogrammetry and Remote Sensing Group, ETH Zurich},
  city={Zurich}, postcode={8093}, country={Switzerland}}

\affiliation[2]{organization={Forest Resources Management Group, ETH Zurich},
  city={Zurich}, postcode={8093}, country={Switzerland}}

\affiliation[3]{organization={ETH Swiss Geo Lab},
  addressline={D4 Business Village, Platz 1a},
  city={Root}, postcode={6039}, country={Switzerland}}

\affiliation[4]{organization={Ecosystem Ecology Group, Swiss Federal Research Institute WSL},
  addressline={Zürcherstrasse 111},
  city={Birmensdorf}, postcode={8902}, country={Switzerland}}

\affiliation[5]{organization={Forest Resources and Management, Swiss Federal Institute for Forest, Snow and Landscape Research WSL},
  city={Birmensdorf}, postcode={8903}, country={Switzerland}}

\cortext[1]{Corresponding author}

\begin{abstract}
\noindent Tree-species composition shapes forest productivity and how forests
respond to climate disturbances such as drought, windthrow and insect outbreaks. Yet most
national forest maps assign each pixel a single dominant species and cannot
show the share of accompanying species. We mapped the basal-area shares of ten
main tree species and a residual Other class across Switzerland at 30~m
resolution. A deep-learning model combining 0.1~m aerial RGB imagery,
airborne laser scanning (ALS), annual AlphaEarth satellite embeddings, and
climate variables was trained on basal-area observations from 3,992 Swiss
National Forest Inventory (NFI) sites. In spatial cross-validation, the model
reached a macro-$R^2$ of 0.485 across the 11 classes. On an independent
forest-reserve dataset, the map reached a macro-$R^2$ of 0.552 for individual
30~m pixels and 0.733 for plots, and identified the dominant class in 73.2\%
of pixels and 81.1\% of plots. The map shows maple and ash as widespread minor
components that dominant-class maps omit. It also reproduces the elevational
sequence from broadleaved lowland stands through spruce to larch and Arolla
pine near the treeline. By resolving the species mixture within each 30~m
pixel, the map complements field inventory statistics with
spatially continuous information that can support forest monitoring, the
management of mixed stands, and climate-adaptation planning across
Switzerland.

\end{abstract}


\maketitle

\section{Introduction}
\label{sec:introduction}

Accurate, spatially explicit information on tree-species composition is
essential for understanding how forests function and respond to
disturbances. Tree diversity is positively associated with forest
productivity \citep{liang2016positive}, and species differ in their
susceptibility to drought, windthrow and insect outbreaks
\citep{anderegg2020climate,senf2021mapping}. Forest services need
composition information to assess current forest condition and to select
species for climate-adapted regeneration
\citep{fassnacht2016review,strauss2025forestreport}. In mixed forests, this
information goes beyond the dominant species. In other words, stands with the same dominant species can differ widely in the number and proportions of accompanying
species. Large-area composition maps therefore need to quantify the shares
of all species in a stand, including the accompanying ones
\citep{bolyn2022mapping,klehr2025fractions}.

Remote sensing (RS) has made large-area tree-species mapping increasingly
feasible by combining wall-to-wall image coverage with forest-inventory
observations. Satellite time series record seasonal differences among
species and provide repeated observations over large, environmentally
diverse regions \citep{blickensdorfer2024national,breidenbach2021norway}.
Combined with national inventory data, they support products that
distinguish multiple species or species groups, including a 10~m map of
the dominant species in Switzerland \citep{koch2024switzerland}. Most of these products,
however, assign a single class to each pixel. At resolutions of 10--30~m,
one pixel can cover several crowns and species, so a single class leaves
the within-pixel composition unresolved \citep{fassnacht2016review}. More
accurate classifiers make the dominant label more reliable, but they
still do not quantify the shares of secondary species.
Mixed stands also complicate the construction of reference data.
Restricting training samples to homogeneous or nearly pure stands
simplifies the link between image and label, but it underrepresents mixed
conditions during model development. Evaluating on pure stands alone then
gives an incomplete picture of performance where mixtures are common
\citep{blickensdorfer2024national}.

Continuous composition targets avoid this simplification by describing
each reference unit as a vector of species proportions. Such targets have
been learned from inventory data of mixed stands with spectral--spatial
deep learning \citep{bolyn2022mapping}, from synthetic mixtures of pure
Sentinel-2 time series \citep{klehr2025fractions}, from airborne laser
scanning \citep{murray2024estimating}, and from combinations of lidar,
airborne spectral and satellite data
\citep{orka2021largearea,candelasbielza2024predicting,cao2026m3fnet}.
All of these studies cover a single region
(Table~\ref{tab:species-mapping-studies}), which
spans a limited range of climate, elevation and species pools, so the
relationship between imagery and composition is learned and tested under
similar environmental conditions \citep{blickensdorfer2024national}.
In contrast, Switzerland is a demanding test case: its
forests extend from the lowlands of the Plateau to the treeline in the
Alps, and their composition shifts from broadleaved and mixed stands at
low elevations to coniferous stands at high elevations
\citep{strauss2025forestreport,wuest2020lidar}.

Very-high-resolution aerial imagery provides evidence that freely available
satellite pixels cannot. At 0.1~m resolution, individual crowns, their texture and their
arrangement are visible, and such imagery has supported individual-tree
detection and species identification in Swiss forests
\citep{beloiu2023individual}. Switzerland acquires such imagery for its
entire territory on a roughly three-year cycle
\citep{swisstopo2026swissimage}. Using it for composition mapping, however,
poses a different learning problem. Inventory plots record the species and
diameter of individual trees, but these records are not matched to crowns
in the image, so a model must relate the appearance of a whole canopy patch
to the plot-level composition. Self-supervised models pretrained on large
unlabeled image collections provide features that transfer to forest
imagery with few labels \citep{simeoni2025dinov3}. Such features have
supported canopy height mapping \citep{brandt2026chmv2}, but they have not
yet been tested for estimating species composition from inventory plots.

Other observations, such as satellite time series, airborne laser
scanning, and climate data, may add what a single aerial acquisition misses.
Species that look alike on one date can differ in their seasonal
trajectories \citep{cloutier2024phenology}, and annual satellite embeddings
summarize such trajectories from several sensors
\citep{brown2025alphaearth,ball2026gfm}. Airborne laser scanning describes
canopy structure \citep{murray2024estimating}, and climate variables
describe where each species can grow \citep{blickensdorfer2024national}.
Combining these sources with aerial imagery is not straightforward. They
differ in spatial support, from 0.1~m image pixels to 10~m embeddings and
1~km climate grids, and in acquisition date. With few training plots, a
model may also rely on covariates that mainly encode location, such as
climate, instead of on what the imagery shows. Fusing satellite imagery and
lidar has improved composition estimates
\citep{candelasbielza2024predicting,cao2026m3fnet}, but most of this work
did not include very-high-resolution aerial imagery, so little is known
about what these sources add to it.

\begin{table}[pos=htbp]
\caption{Large-area studies and products mapping tree species or leaf type.
S-1/S-2 = Sentinel-1/2; ALS = airborne laser scanning; HSI = hyperspectral
imagery; AEF = AlphaEarth Foundations satellite embeddings.}
\label{tab:species-mapping-studies}
\setlength{\tabcolsep}{3pt}
\renewcommand{\arraystretch}{1.15}
\renewcommand{\tabularxcolumn}[1]{>{\raggedright\arraybackslash}m{#1}}
\small
\begin{tabularx}{\textwidth}{@{}>{\raggedright\arraybackslash}m{4.3cm}
                              >{\raggedright\arraybackslash}X
                              >{\raggedright\arraybackslash}m{4.9cm}
                              >{\raggedright\arraybackslash}m{2.2cm}}
\toprule
\textbf{Study/product} & \textbf{Prediction target} & \textbf{Predictor modalities} & \textbf{Scale} \\
\midrule
\multicolumn{4}{@{}l}{\textit{Categorical}} \\
\cmidrule{1-4}
\citet{francini2024forest} & 7 species groups & S-2 & Netherlands \\
\citet{mu2025forestformer} & 8 species/species groups & S-2 & Germany \\
\citet{waser2017walltowall} & 2 leaf types; broadleaf probability & Aerial multispectral + terrain & Switzerland \\
\citet{blickensdorfer2024national} & 11 species/species groups & S-1 + S-2 + environment & Germany \\
\citet{koch2024switzerland} & 15 species & S-2 + terrain & Switzerland \\
\citet{waser2025foresttype} & 2 leaf types; broadleaf probability & S-1 + S-2 + terrain & Switzerland \\
\citet{breidenbach2021norway} & 3 species/species groups & S-2 + terrain & Norway \\
\citet{abdi2026sweden} & 8 (7 species, 1 Other) & S-1 + S-2 + terrain + canopy height & Regional \\
\addlinespace[3pt]
\multicolumn{4}{@{}l}{\textit{Continuous}} \\
\cmidrule{1-4}
\citet{bolyn2022mapping} & 9 (6 genera, 2 species, 1 Other) & S-2 & Regional \\
\citet{murray2024estimating} & 9 (7 species, 2 genera) & ALS & Regional \\
\citet{klehr2025fractions} & 10 (9 species, 1 Other) & S-2 & Regional \\
\citet{orka2021largearea} & 3 groups & ALS + airborne HSI & Regional \\
\citet{cao2026m3fnet} & 9 (7 species, 2 genera) & S-2 + ALS & Regional \\
\citet{candelasbielza2024predicting} & 3 groups & ALS + aerial multispectral + S-2 & Regional \\
\addlinespace[3pt]
\textbf{Ours} & \textbf{11 (10 main tree species, 1 Other)} & \textbf{RGB + ALS + AEF + climate} & \textbf{Switzerland} \\
\bottomrule
\end{tabularx}
\end{table}
\renewcommand{\tabularxcolumn}[1]{p{#1}}
\renewcommand{\arraystretch}{1.35}

Here, we aim to develop a national, spatially continuous map of tree-species composition for Switzerland, representing the basal-area shares of 10 main tree species and an Other class rather than assigning each location to a single dominant species. We specifically ask: i) Within an RGB-anchored multimodal framework, what predictive information does each remote-sensing and environmental data source contribute to tree-species composition mapping? ii) How accurately does the national map estimate tree-species composition across independent spatial scales and datasets? iii) What ecological patterns of tree-species composition emerge across Switzerland’s elevational gradient? We hypothesize that integrating multimodal data enables accurate mapping of tree-species composition beyond inventory sites while preserving ecologically meaningful spatial patterns. We address these questions by combining basal-area observations from 3,992 Swiss National Forest Inventory sites with high-resolution aerial imagery, airborne laser scanning, satellite embeddings, and climate data in a multimodal deep-learning framework. We evaluate the contribution of the different data sources using spatial cross-validation, generate a 30 m national composition map, and independently evaluate it using Swiss Forest Reserves and regional NFI4 statistics. Continuous composition mapping provides a new basis for monitoring forest diversity and change.

\section{Data}
\label{sec:methods}

\subsection{Study area and Swiss National Forest Inventory}
\label{sec:study-area}
\label{sec:nfi}

\begin{figure}[pos=htbp]
\centering
{\raggedright\panellabelfont (a)\par}\vspace{2pt}
\includegraphics[width=\textwidth]{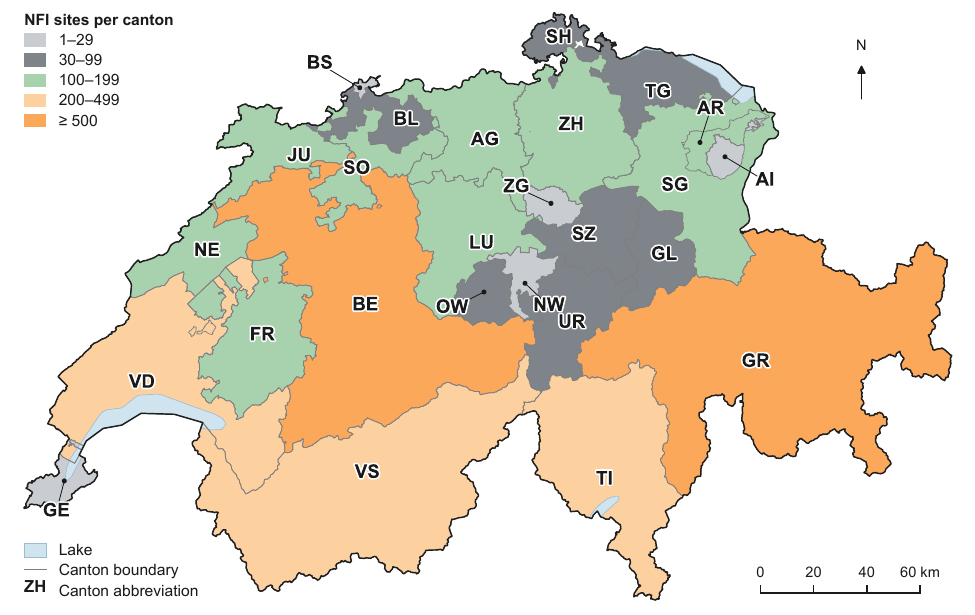}
\par\bigskip
\setlength{\panelht}{\dimexpr\textwidth-14pt\relax}
\setlength{\panelht}{0.3389\panelht}
\noindent\makebox[\textwidth]{%
  \abovepanel{\panelht}{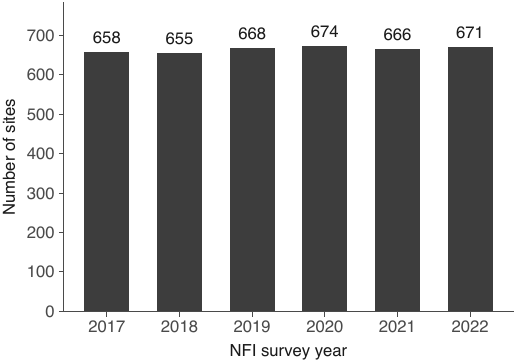}{(b)}%
  \hfill
  \abovepanel{\panelht}{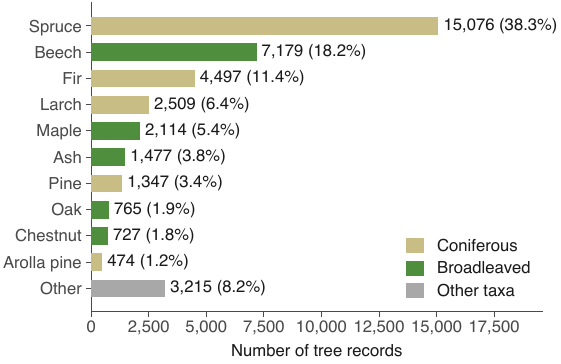}{(c)}%
}
\caption{Overview of the Swiss National Forest Inventory (NFI) sites used in
our study. (a) Number of retained NFI sites in each of Switzerland's 26 cantons,
in five classes (3,992 sites in total); cantons are labeled with their
official two-letter abbreviations. (b) Number of retained sites by NFI survey year (2017--2022). (c)
Number of retained tree-level observations assigned to each of the 10 Swiss
NFI main tree species and the residual Other class, colored by leaf type. Here and in all other maps of this article and its Supplementary Material, map lines delineate study areas and do not necessarily depict accepted national boundaries.}
\label{fig:study-area}
\end{figure}

Switzerland (41,285~km$^2$) spans pronounced elevational and climatic
gradients, from the low-lying Plateau and Jura foothills to the high Alps
(Fig.~\ref{fig:study-area}a).
These gradients are important drivers of woody species distributions across
the country \citep{wuest2020lidar}. Forests cover 1.3 million hectares, or
32\% of the land. Nationally, 61\% of the forested area is coniferous and
39\% broadleaved \citep{strauss2025forestreport}. Forest composition,
however, varies strongly among regions: coniferous, spruce-dominated stands
predominate in parts of the Alps, whereas broadleaved and
mixed forests occur more frequently at lower elevations and in the warmer
southern regions \citep{strauss2025forestreport,wuest2020lidar}.

Reference data were derived from the Swiss National Forest Inventory (NFI),
a systematic national survey on a 1.4~$\times$~1.4~km sampling grid
\citep{wsl2023nfidata,fischertraub2019nfi,strauss2025forestreport}.
Field crews record species identity, diameter at breast height (DBH,
measured at 1.3~m), and canopy stratum for individual trees. Stems with DBH
12--35.9~cm are sampled within 200~m$^2$, and those with DBH
$\geq$36~cm within 500~m$^2$
\citep{fischertraub2019nfi,duggelin2020nfi5}. The corresponding
per-hectare expansion factors, 50 and 20, were applied when aggregating
basal area to plots.

\subsection{Reference label construction}
\label{sec:labels}

The prediction taxonomy follows the Swiss NFI main-tree-species grouping
\citep{fischertraub2019nfi,strauss2025forestreport}. Ten species or genera
are mapped separately: spruce, fir, pine, larch, Arolla pine, beech, maple,
ash, oak and chestnut. The remaining 48 retained taxa, both broadleaved and
coniferous, are pooled into an Other class (Table~\ref{tab:species-groups}); example
sites of the ten species are shown in Fig.~\ref{fig:cv-folds-map}b.
From 63,969 tree records at 5,998 NFI sites, we retained observations from
2017 onward, when SWISSIMAGE acquisitions become available for download
\citep{swisstopo2026swissimage}. Upper- and middle-stratum trees were
retained, and understory trees were excluded because they are largely
hidden in nadir-viewing imagery. Canopy strata follow the NFI5 protocol
\citep{duggelin2020nfi5}. We also excluded sites affected by
stand-replacing disturbance between the field survey and RS acquisitions,
sites whose SWISSIMAGE imagery was degraded, mostly by terrain or cloud
shadow, and sites with incomplete multimodal records. The resulting dataset
contains 39,380 tree records at 3,992 sites surveyed in 2017--2022
(Fig.~\ref{fig:study-area}b,c).

We express species composition as shares of basal area. Basal area weights
each tree by its size and is derived from the DBH recorded for every
measured tree. It is also reported in the published NFI statistics
\citep{abegg2023nfitables}, against which the map is compared in
Section~\ref{sec:res-nfi4}. For tree $j$ at site $i$, the expanded basal
area is $a_{ij}=w_{ij}\pi(\mathrm{DBH}_{ij}/200)^2$ in m$^2$~ha$^{-1}$,
where DBH is in centimeters and $w_{ij}$ is the expansion factor defined in
Section~\ref{sec:nfi}. If $z_{ij}$ denotes the mapped class of the tree,
the class-specific basal area and composition are

\begin{equation}
A_{ik}=\sum_{j:z_{ij}=k}a_{ij},
\qquad
y_{ik}=\frac{A_{ik}}{\sum_{k'=1}^{11}A_{ik'}},
\quad k=1,\ldots,11.
\label{eq:species}
\end{equation}

The vector $\mathbf{y}_i$ sums to one. It gives the share of each class in
the expanded basal area of the upper and middle strata. Two auxiliary
targets summarize stand purity and the broadleaved--coniferous composition
(Section~\ref{sec:model-heads}).

\subsection{Remote sensing and environmental data}
\label{sec:sources}
\label{sec:harmonisation}

Four inputs were extracted around each NFI site coordinate: aerial RGB
imagery, airborne laser scanning (ALS) point clouds, annual satellite
embeddings, and climate variables (Table~\ref{tab:datasets}). Aerial imagery was obtained from
SWISSIMAGE, the 0.1~m RGB orthomosaic of the Swiss Federal Office of
Topography (swisstopo) \citep{swisstopo2026swissimage}. For each site, we
cropped a 28~m~$\times$~28~m square (280~$\times$~280 pixels) centered on
the site coordinate. This extent slightly exceeds the 25.2~m diameter of the
outer NFI sample circle, so that the crop accommodates coordinate
uncertainty and crowns extending beyond the plot boundary. SWISSIMAGE is
renewed on a roughly three-year cycle, and acquisitions differ in season.
Leaf-on acquisitions were selected following \citet{ferrari2026deadwood}.
Canopy structure was represented by swissSURFACE3D, the nationwide ALS
point cloud of swisstopo, with a mean density of about
15--20~points~m$^{-2}$ \citep{swisstopo2026swisssurface3d}. Point clouds
were cropped to a 14~m-radius circle centered on each site and inscribed in
the RGB crop. Only the $X$, $Y$, and $Z$ coordinates and the point
classification were retained. Satellite information was represented by
annual, 64-dimensional AlphaEarth embeddings at 10~m resolution
\citep{brown2025alphaearth}, from which a 3~$\times$~3-pixel patch centered
on each site was extracted for the NFI survey year. Climate was
represented by the 19 CHELSA BIOCLIM+ variables for the 1981--2010
climatology at approximately 1~km resolution \citep{brun2022chelsa}, read
at each site's coordinate.

\begin{table}[pos=htbp]
\caption{Input data modalities. ALS = airborne laser scanning.}
\label{tab:datasets}
\setlength{\tabcolsep}{3pt}
\renewcommand{\arraystretch}{1.15}
\renewcommand{\tabularxcolumn}[1]{>{\raggedright\arraybackslash}m{#1}}
\small
\begin{tabularx}{\textwidth}{@{}>{\raggedright\arraybackslash}m{3.4cm}
                              >{\raggedright\arraybackslash}m{3.8cm}
                              >{\raggedright\arraybackslash}m{3.2cm}
                              >{\raggedright\arraybackslash}X}
\toprule
\textbf{Modality} & \textbf{Dataset} & \textbf{Spatial resolution} & \textbf{Acquisition time} \\
\midrule
Aerial RGB & SWISSIMAGE & 0.1~m & Leaf-on acquisition nearest the NFI survey year \\
ALS & swissSURFACE3D & 15--20 points~m$^{-2}$ & Single available acquisition \\
Satellite embeddings & AlphaEarth Foundations & 10~m & NFI survey year \\
Climate & CHELSA BIOCLIM+ & ${\sim}$1~km & 1981--2010 long-term mean \\
\bottomrule
\end{tabularx}
\end{table}

\subsection{Forest reserve data}
\label{sec:reserve-data}

\begin{figure}[pos=htbp]
\centering
\includegraphics[width=\textwidth]{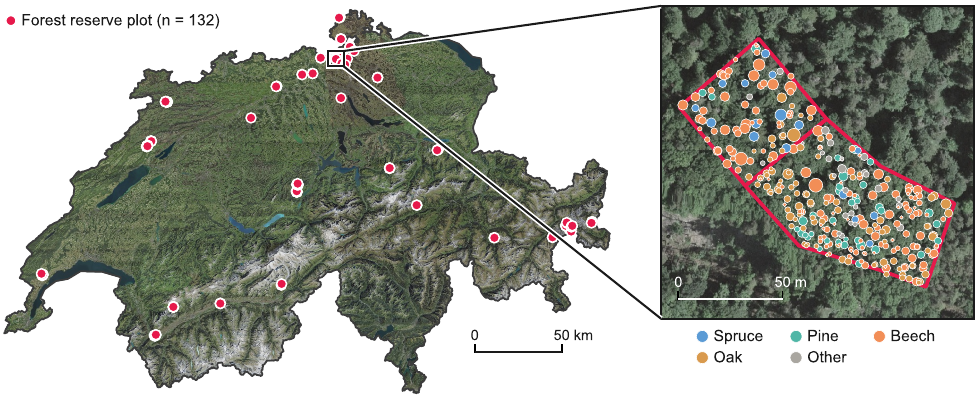}
\caption{Forest reserve validation data. Left: locations of the 132 plots
of the Swiss Forest Reserves Monitoring network used for independent
validation. Right: an example plot polygon (red outline), with circles
marking the retained trees colored by mapped class and sized by DBH.}
\label{fig:reserve-data}
\end{figure}

For independent validation, we used the Swiss Forest Reserves Monitoring
network \citep{hobi2020reserves}, which was established to study the
long-term dynamics of unmanaged forests. Each reserve plot is a permanent
polygon in which every tree is recorded with its species, DBH, canopy
stratum, and position. The plots range from 759 to 34,704~m$^2$, with a
median of about 4,689~m$^2$, so most of them span several 30~m pixels
(Fig.~\ref{fig:reserve-data}). For each plot, we used the most recent
census and retained living upper- and middle-stratum trees with
DBH~$\geq$~12~cm, matching the label definition of
Section~\ref{sec:labels}. Species were assigned to the 11 classes with the
same crosswalk as the NFI labels, and composition was computed as
basal-area shares (Eq.~\ref{eq:species}). The validation dataset contains
132 plots in 35 reserves.

\section{Methods}
\label{sec:model}

\begin{figure}[pos=htbp]
\centering
    \includegraphics[width=\textwidth]{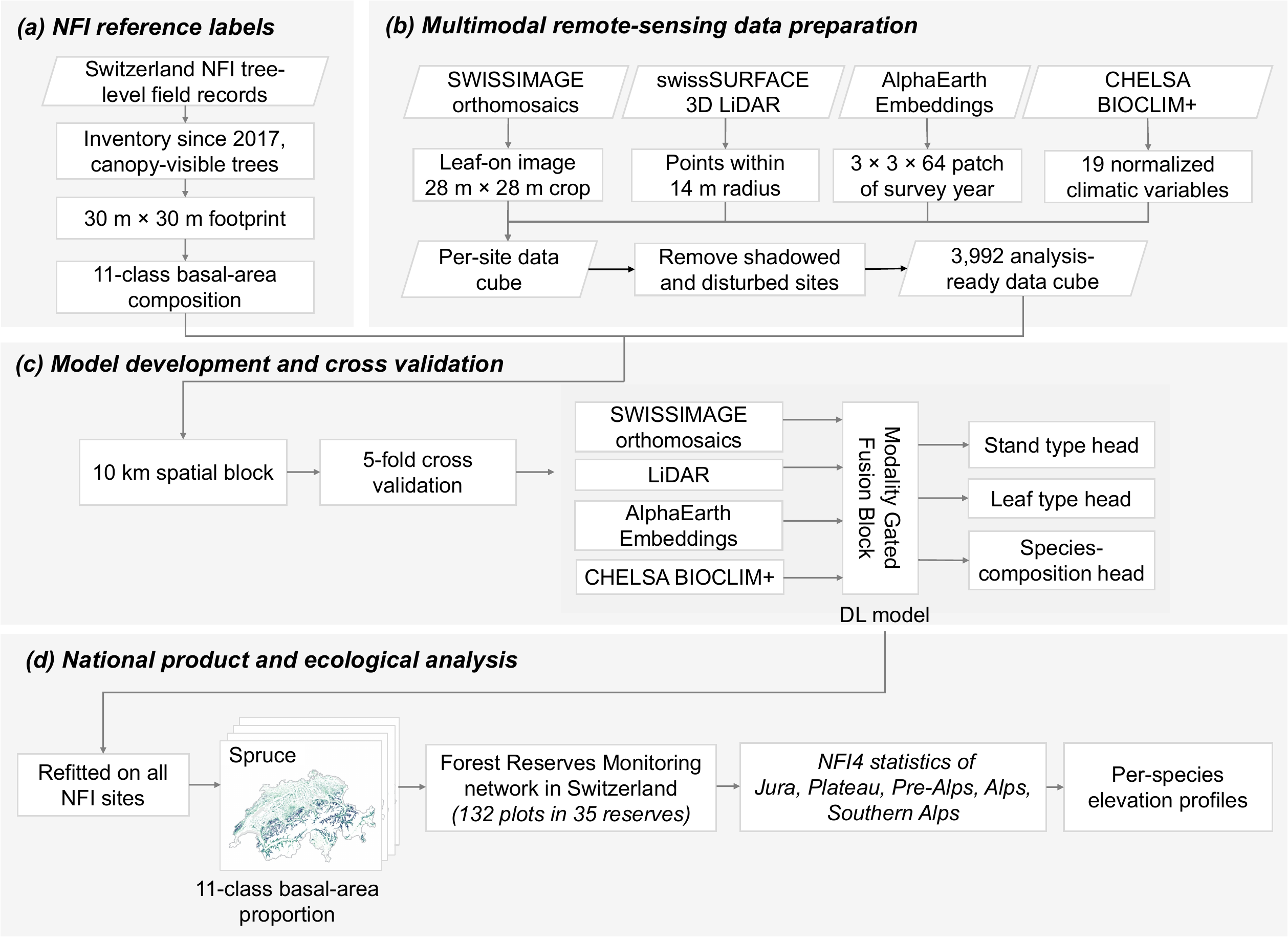}
\caption{Workflow of this study. (a) Construction of site-level
basal-area compositions from NFI tree records. (b) Extraction and screening
of the four input modalities. (c) Model training and five-fold spatial
block cross-validation. (d) National mapping with the refitted model and
independent evaluation.}
\label{fig:pipeline}
\end{figure}

Figure~\ref{fig:pipeline} summarizes the workflow. (a) NFI tree records
were converted into site-level basal-area compositions
(Section~\ref{sec:labels}). (b) The four input modalities were extracted at
each site, and sites with degraded imagery or stand-replacing disturbance
were removed (Sections~\ref{sec:labels} and~\ref{sec:sources}). (c) A
multimodal network was trained and evaluated with five-fold spatial block
cross-validation (Sections~\ref{sec:architecture} and~\ref{sec:training}).
(d) The selected model was refitted on all NFI sites and applied across the
Swiss forest at 30~m, and the map was evaluated against forest-reserve
plots, NFI4 regional statistics, and elevational patterns
(Sections~\ref{sec:inference-data} and~\ref{sec:accuracy}).

\subsection{Model architecture}
\label{sec:architecture}

\begin{figure}[pos=htbp]
\centering
\includegraphics[width=0.8\textwidth]{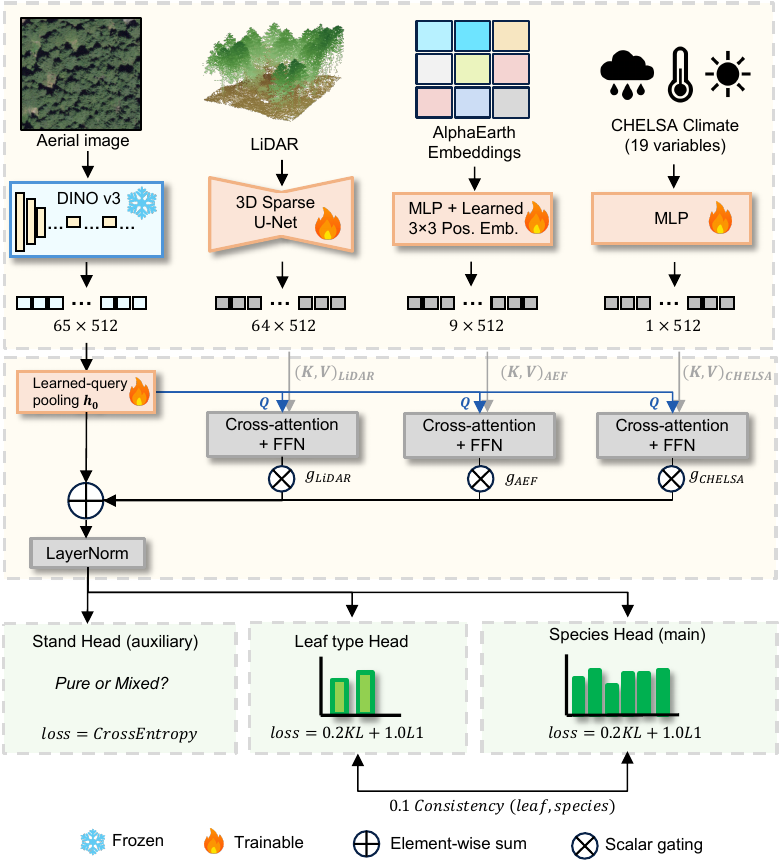}
\caption{Overview of the multimodal architecture. Modality-specific encoders
project each input into a token set in a shared embedding space: a frozen DINOv3
ViT-H+/16 for the SWISSIMAGE crop, a sparse 3D U-Net with learned queries
for the ALS point cloud, and multilayer perceptrons for the AlphaEarth patch
and the CHELSA variables (Section~\ref{sec:model-encoders}). A learned query
pools the aerial features into the anchor $\mathbf{h}_0$; the other
modalities enter as residual corrections $\boldsymbol{\Delta}_m$, each scaled
by a learned gate $g_m$ (Section~\ref{sec:model-fusion}). Three heads predict
stand type, leaf type, and species composition, with the loss terms of
Section~\ref{sec:model-heads}.}
\label{fig:architecture}
\end{figure}

Supervision is provided at the site level. Each training sample pairs the
30~m~$\times$~30~m footprint with its observed composition, so the model
learns to aggregate evidence over the whole footprint. The architecture has
three stages (Fig.~\ref{fig:architecture}). Modality-specific encoders
project each input into a shared $D$-dimensional embedding space; a gated
fusion module combines these embeddings into a single site representation
anchored on the aerial imagery; and three heads predict the species
composition and two auxiliary targets, stand type and leaf type
(Section~\ref{sec:model-heads}).

\subsubsection{Modality encoders}
\label{sec:model-encoders}

Aerial imagery is encoded by a frozen DINOv3 ViT-H+/16
\citep{simeoni2025dinov3}, pretrained self-supervised on LVD-1689M. To
combine local texture with higher-level features, we read out four evenly
spaced blocks, $\{7,15,23,31\}$, combine them with learned softmax weights,
and keep the class token together with the patch tokens pooled to an
$8\times8$ grid. The ALS point cloud is encoded with a randomly-initialized sparse 3D U-Net
that follows the backbone architecture of ForestFormer3D
\citep{xiang2025forestformer3d}. Sixty-four learned queries
then pool the voxel features by attention into a fixed-size token set,
independent of the number of occupied voxels at a site. AlphaEarth embeddings \citep{brown2025alphaearth} already summarize a year
of multi-sensor satellite observations. A shared two-layer MLP with a
positional embedding projects each of the $3\times3$ pixels into the
embedding space. The 19 CHELSA variables are encoded by a second two-layer
MLP into a single token, as in GlobalGeoTree \citep{mu2025globalgeotree}.

\subsubsection{RGB-anchored gated fusion}
\label{sec:model-fusion}

The modalities differ widely in spatial support, from 0.1~m aerial
imagery to 1~km climate data, and in how directly they relate to the
species within a site \citep{qin2025multibranch,mu2025globalgeotree}. Only
the aerial imagery resolves individual crowns within the footprint, so we
use it as an anchor and let each other modality contribute a gated
correction. Let $\mathbf{T}_m$ denote the token set produced by the encoder
of modality $m$, and $\mathbf{M}_m$ a binary mask that flags tokens derived
from missing or quality-masked observations. A learned query $\mathbf{q}$
first pools the aerial tokens into the anchor,

\begin{equation}
\mathbf{h}_0=\mathrm{LN}\!\left(\mathrm{Attn}\!\left(\mathbf{q},\,
\mathbf{T}_{\mathrm{rgb}},\,\mathbf{T}_{\mathrm{rgb}}\right)\right)
\ \in\ \mathbb{R}^{D}.
\label{eq:anchor}
\end{equation}

Each auxiliary modality $m\in\{\mathrm{ALS},\ \mathrm{embedding},\
\mathrm{climate}\}$ then reads its own token set with $\mathbf{h}_0$ as the
query,

\begin{equation}
\mathbf{c}_m=\mathrm{Attn}\!\left(\mathbf{W}_q\mathbf{h}_0,\
\mathbf{T}_m,\ \mathbf{T}_m;\ \mathbf{M}_m\right),
\qquad
\boldsymbol{\Delta}_m=\mathbf{c}_m+\mathrm{FFN}(\mathbf{c}_m),
\label{eq:correction}
\end{equation}

where $\mathrm{Attn}(\cdot)$ is multi-head attention with an optional key
mask, $\mathrm{LN}$ is layer normalization, $\mathrm{FFN}$ is a two-layer
feed-forward network, and $\mathbf{W}_q$ is a learned projection. Each
auxiliary modality attends only to its own tokens, so the token sets can
differ in size across modalities. Each correction is admitted through a gate predicted per site,

\begin{equation}
g_m=\sigma\!\left(\phi_m\!\left(\left[\mathbf{h}_0;\
\bar{\mathbf{t}}_m\right]\right)\right)\cdot
\mathbf{1}\!\left[\textstyle\sum_i \bigl(1-M_m^{(i)}\bigr)>0\right],
\qquad
\mathbf{h}=\mathrm{LN}\!\left(\mathbf{h}_0+\sum_m g_m\,
\boldsymbol{\Delta}_m\right),
\label{eq:gate}
\end{equation}

where $\sigma$ is the logistic function, $\phi_m$ a two-layer perceptron, and
$\bar{\mathbf{t}}_m$ the masked mean of $\mathbf{T}_m$; the indicator closes
the gate when a modality is missing. The last layer of each $\phi_m$ is
initialized with zero weights and bias $-4$, so that $g_m\approx0.018$ at the
start and training begins close to an imagery-only model. During
training, each auxiliary modality was dropped with probability 0.1 per site
by forcing $g_m=0$ \citep{neverova2016moddrop}, so that the model remains
usable when an input is missing.

\subsubsection{Prediction heads and loss}
\label{sec:model-heads}

Besides the 11-part composition, the model predicts two coarser targets
derived from the same basal-area observations. Stand type marks a site as
pure when its largest class share reaches 0.80 and as mixed otherwise.
Leaf type is the pair of broadleaved and coniferous shares of the expanded
basal area. It is computed from the original NFI taxa rather than from the
11 mapped classes, so the broadleaved and coniferous taxa within Other are
assigned to their own leaf type. Both
targets are easier to recognize than individual species and give the
shared representation additional supervision. The three targets are
predicted in parallel rather than hierarchically, so that an error in a
coarse target cannot propagate to the species predictions, as it can in
hierarchical classification
\citep{li2025enmap,ball2026gfm,luo2026cognition,turkoglu2021cropmapping}.
Three heads read $\mathbf{h}$: a linear layer for stand type, a linear
layer for leaf type, and a two-layer perceptron ($D\to256\to11$) for
composition, each followed by a softmax.

Stand type is supervised with cross-entropy, $\mathcal{L}_{\mathrm{stand}}$.
The two compositional targets combine a Kullback--Leibler divergence, which
matches the shape of the distribution, with an $L_1$ term, which penalizes
absolute proportion errors and is less dominated by abundant classes,

\begin{equation}
\mathcal{L}_{\mathrm{comp}}(\hat{\mathbf{p}},\mathbf{y})=
\lambda_{\mathrm{KL}}\sum_k y_k\log\frac{y_k}{\hat{p}_k}
+\lambda_{L_1}\sum_k\left|\hat{p}_k-y_k\right|,
\label{eq:comp-loss}
\end{equation}

with $\lambda_{\mathrm{KL}}=0.2$ and $\lambda_{L_1}=1.0$. We denote this
loss for the leaf-type and composition targets by
$\mathcal{L}_{\mathrm{leaf}}$ and $\mathcal{L}_{\mathrm{species}}$. A
consistency term $\mathcal{L}_{\mathrm{cons}}$ couples the leaf-type and
composition heads. Summing the predicted shares of the broadleaved main
classes gives a lower bound $B_{\mathrm{known}}$ on the broadleaved share,
and Other can add at most $\hat{p}_{\mathrm{Other}}$ to it.
$\mathcal{L}_{\mathrm{cons}}$ is a hinge penalty that is zero when the
predicted broadleaved share $\hat{p}_{\mathrm{leaf},B}$ lies within
$[B_{\mathrm{known}},\,B_{\mathrm{known}}+\hat{p}_{\mathrm{Other}}]$ and
grows linearly outside this interval. The total loss is

\begin{equation}
\mathcal{L}=\mathcal{L}_{\mathrm{stand}}+\mathcal{L}_{\mathrm{leaf}}
+\mathcal{L}_{\mathrm{species}}
+\lambda_{\mathrm{cons}}\mathcal{L}_{\mathrm{cons}},
\qquad \lambda_{\mathrm{cons}}=0.1.
\label{eq:total-loss}
\end{equation}

\subsection{Spatial cross-validation and training}
\label{sec:training}
\label{sec:splits}

\begin{figure}[pos=htbp]
\centering
\includegraphics[width=0.98\textwidth]{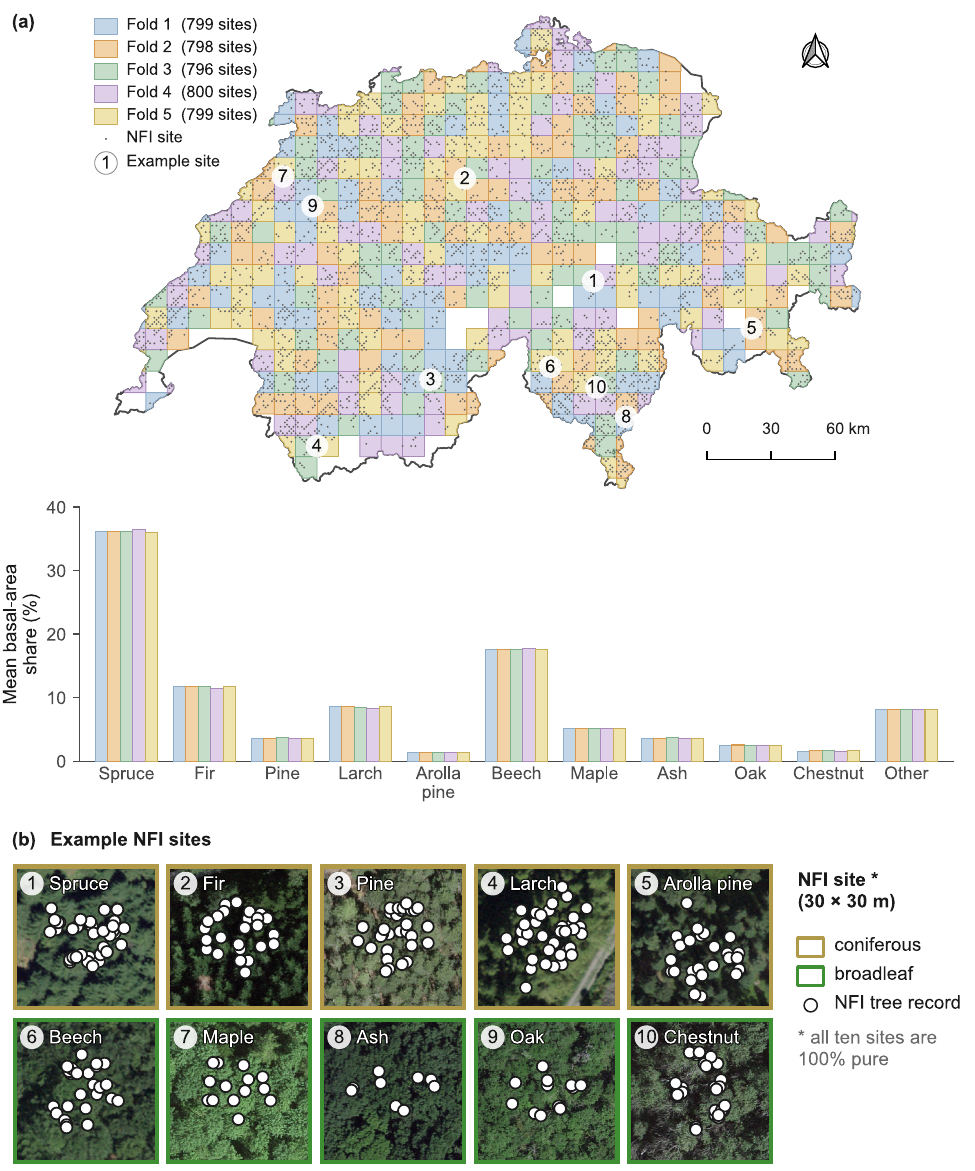}
\caption{Spatial cross-validation design and example reference sites.
(a) The 3,992 retained NFI sites (gray dots) were grouped into 10~km spatial
blocks, and each block was assigned to one of the five cross-validation
folds; the bar chart gives the mean site-level basal-area share of each
mapped class in the five folds. Numbered markers (1--10) locate the example
sites in (b). (b) One pure 30~$\times$~30~m site for each of the ten main
tree species, with white circles marking the recorded NFI trees and the
frame color indicating leaf type.}
\label{fig:cv-folds-map}
\end{figure}

To evaluate prediction at unseen locations while accounting for spatial
autocorrelation \citep{roberts2017crossvalidation}, we grouped the 3,992
sites into 10~km spatial blocks and assigned entire blocks to five folds
\citep{abdi2026sweden}. A greedy assignment balanced the fold sizes and the
mean basal-area share of each of the 11 classes across folds, a form of
spatially stratified block cross-validation \citep{valavi2019blockcv}. The
folds were rotated through training and evaluation
(Fig.~\ref{fig:cv-folds-map}a), and each site was predicted by the model
for which it was held out, so that all 3,992 sites enter the evaluation.
Each model was trained for 80 epochs with AdamW (learning rate
$1\times10^{-4}$, weight decay 0.05, cosine decay to $1\times10^{-6}$) and
a batch size of 4, with an exponential moving average of the weights
(decay 0.999). Sites containing rare classes were oversampled; further
settings are given in Supplementary Section~\ref{sec:appendix-training}.

We compared the model with three baselines: a random forest, a ResNet-50
image model, and a multilayer perceptron on AlphaEarth features
\citep{tong2026invariant}. We also replaced the ViT-H+/16 encoder with
ViT-L/16 encoders pretrained on LVD-1689M and SAT-493M. Modality ablations
removed one or more inputs and retrained the model under the same
protocol. The configuration with the highest cross-validated macro-$R^2$
was selected.

\subsection{National-scale mapping}
\label{sec:inference-data}

After model selection, the selected configuration was refitted on all 3,992
sites to obtain the production model. It was applied on a 30~m grid within
the forest mask of \citet{ferrari2026deadwood}. The 30~m cell matches the
spatial support of the reference labels: at 99.0\% of the sites, all
retained tree positions fall within a centered 30~m cell, compared with
65.5\% for a 20~m cell. At each grid cell, we used the most recent leaf-on
SWISSIMAGE acquisition and the AlphaEarth embedding of the same year,
together with swissSURFACE3D and the CHELSA climatology. The reference year
of the map therefore varies between regions, from 2018 to 2025
(Fig.~\ref{fig:inference-year}).

\subsection{Accuracy assessment}
\label{sec:accuracy}

Following \citet{nelson2021mapaccuracy}, we assessed the product from three
perspectives: site-specific accuracy, non-site-specific accuracy, and
spatial-distribution accuracy. For each class, we computed the mean
absolute error (MAE), the root mean squared error (RMSE), and the
coefficient of determination ($R^2$) between observed and predicted shares.
We averaged them over the classes (macro-average), so that every class
contributes equally regardless of its abundance. Each composition vector
was also converted to its dominant class, for which we report overall
accuracy and macro-$F_1$. Cross-validation metrics are reported as the
mean~$\pm$~standard deviation over the five held-out folds.

Site-specific accuracy was assessed with the cross-validated predictions at
the 3,992 NFI sites and with the forest reserve data
(Section~\ref{sec:reserve-data}) at two supports. At plot level, the
prediction was the intersection-area-weighted mean of all 30~m pixels
overlapping the plot polygon. At pixel level, we retained pixels at least
80\% covered by a plot and containing at least 10 trees, and compared each
with its own prediction (414 pixels from 111 plots in 34 reserves).
Chestnut occurs in only one reserve plot, so its $R^2$ cannot be
evaluated, and reserve $R^2$ is averaged over the remaining ten classes. Non-site-specific accuracy was assessed against the published NFI4
basal-area shares of the five production regions
\citep{abegg2023nfitables}, for which the predicted shares were averaged
over the forest pixels of each region. Spatial-distribution accuracy was
assessed along the elevation gradient. Terrain elevation from swisstopo
swissALTIRegio \citep{swisstopo2026swissaltiregio}, a 10~m model based on
swissALTI3D within Switzerland, was area-averaged onto the 30~m product
grid. Predicted shares were then averaged within 100~m elevation bands, and
only bands containing at least 5,000 forest pixels (4.5~km$^2$) were
analyzed.

\section{Results}
\label{sec:results}

We first report cross-validated model performance
(Section~\ref{sec:res-model}) and the national composition map
(Section~\ref{sec:res-production}). We then compare the map with
independent forest-reserve data and NFI4 regional statistics
(Section~\ref{sec:res-validation}) and describe its composition along the
elevation gradient (Section~\ref{sec:res-elevation}).

\subsection{Model evaluation}
\label{sec:res-model}

\subsubsection{Comparison with baselines and visual backbones}
\label{sec:res-baselines}

For species-composition prediction, the ViT-H+/16 model performed best on
all three metrics (Table~\ref{tab:baselines}). Its composition $R^2$ of
0.485 exceeded the strongest baseline, ResNet-50, by 0.105 and the random
forest by 0.172, and its MAE was 17\% and 42\% lower (0.048 against 0.058
and 0.083). The baselines differed more in dominant-class $F_1$ than in
MAE. The random forest and the AlphaEarth MLP had almost the same MAE
(0.083 and 0.084), but their $F_1$ was only 0.342 and 0.194, against 0.515
for ResNet-50, so both identified the dominant class far less reliably.
Among the DINOv3 backbones, the larger ViT-H+/16 exceeded both ViT-L/16
variants by about 0.04 in $R^2$ and 0.015--0.019 in $F_1$, whereas the two
ViT-L/16 variants, pretrained on different corpora, differed by only 0.002
in $R^2$ and 0.004 in $F_1$. ViT-H+/16 also showed the smallest
fold-to-fold variation in $F_1$ (SD = 0.014). Leaf-type performance varied
less among models, with every configuration reaching $R^2>0.66$.

\begin{table}[pos=htbp]
\caption{Comparison with reference baselines and across frozen DINOv3 backbones. Best value in bold, second best underlined.}
\label{tab:baselines}
\footnotesize
\setlength{\tabcolsep}{3pt}
\renewcommand{\arraystretch}{1.5}
\begin{tabular*}{\textwidth}{@{\extracolsep{\fill}}lllllll@{}}
\toprule
\multirow{2}{*}{\textbf{Model}} & \multirow{2}{*}{\textbf{Backbone}}
 & \multicolumn{2}{c}{\textbf{Leaf level}} & \multicolumn{3}{c}{\textbf{Species level}} \\
\cmidrule(lr){3-4}\cmidrule(l){5-7}
 & & \textbf{MAE} & \textbf{$R^2$} & \textbf{MAE} & \textbf{$R^2$} & \textbf{$F_1$} \\
\midrule
ResNet-50 & ResNet-50, ImageNet & 0.109$\pm$0.004 & 0.757$\pm$0.020 & 0.058$\pm$0.002 & 0.380$\pm$0.043 & 0.515$\pm$0.027 \\
Random Forest & -- & 0.167$\pm$0.004 & 0.669$\pm$0.017 & 0.083$\pm$0.001 & 0.313$\pm$0.012 & 0.342$\pm$0.031 \\
AlphaEarth MLP & -- & 0.149$\pm$0.008 & 0.667$\pm$0.027 & 0.084$\pm$0.002 & 0.157$\pm$0.014 & 0.194$\pm$0.012 \\
\midrule
\multirow{3}{*}{Ours}
 & ViT-L/16, SAT-493M   & 0.103$\pm$0.006 & 0.781$\pm$0.020 & 0.052$\pm$0.002 & \underline{0.446$\pm$0.062} & \underline{0.570$\pm$0.043} \\
 & ViT-L/16, LVD-1689M  & \underline{0.097$\pm$0.004} & \underline{0.794$\pm$0.013} & \underline{0.051$\pm$0.002} & 0.444$\pm$0.063 & 0.566$\pm$0.051 \\
 & ViT-H+/16, LVD-1689M & \textbf{0.093$\pm$0.003} & \textbf{0.803$\pm$0.017} & \textbf{0.048$\pm$0.001} & \textbf{0.485$\pm$0.038} & \textbf{0.585$\pm$0.014} \\
\bottomrule
\end{tabular*}
\end{table}
\renewcommand{\arraystretch}{1.35}

\subsubsection{Contribution of the input modalities}
\label{sec:res-modalities}

Among the auxiliary modalities, AlphaEarth embeddings and CHELSA climate
variables made the largest contributions to composition $R^2$, whereas ALS
provided little additional improvement (Table~\ref{tab:modalities}).
Removing AlphaEarth embeddings reduced $R^2$ from 0.485 to 0.464, and
removing CHELSA reduced it to 0.465; removing ALS reduced it only to 0.483.
RGB imagery alone reached an $R^2$ of 0.452, so the three auxiliary
modalities together added 0.033. MAE and $F_1$ changed much less than
$R^2$: across all six configurations, species-level MAE stayed between
0.048 and 0.051 and $F_1$ between 0.573 and 0.593. The
RGB--AlphaEarth--CHELSA configuration reached the highest $F_1$ (0.593) and
leaf-type $R^2$ (0.814). At leaf level, all configurations without
AlphaEarth had an MAE of 0.099, against 0.093--0.094 with it, suggesting
that the satellite embeddings mainly sharpen the broadleaved--coniferous
distinction. The $R^2$ differences are smaller than the fold-to-fold
standard deviation (0.04--0.07) and should be read as tendencies.

\begin{table}[pos=htbp]
\caption{Leave-one-modality-out and reduced-modality ablations. AEF = AlphaEarth annual embeddings. Best value in bold, second best underlined.}
\label{tab:modalities}
\footnotesize
\setlength{\tabcolsep}{3pt}
\renewcommand{\arraystretch}{1.3}
\begin{tabular*}{\textwidth}{@{\extracolsep{\fill}}lllllllll@{}}
\toprule
\multicolumn{4}{c}{\textbf{Modality}} & \multicolumn{2}{c}{\textbf{Leaf level}} & \multicolumn{3}{c}{\textbf{Species level}} \\
\cmidrule(lr){1-4}\cmidrule(lr){5-6}\cmidrule(l){7-9}
\textbf{RGB} & \textbf{ALS} & \textbf{AEF} & \textbf{CHELSA} & \textbf{MAE} & \textbf{$R^2$} & \textbf{MAE} & \textbf{$R^2$} & \textbf{$F_1$} \\
\midrule
\checkmark & \checkmark & \checkmark & \checkmark & \underline{0.093$\pm$0.003} & 0.803$\pm$0.017 & \textbf{0.048$\pm$0.001} & \textbf{0.485$\pm$0.038} & \underline{0.585$\pm$0.014} \\
\checkmark & \checkmark & \checkmark &            & 0.094$\pm$0.002 & \underline{0.808$\pm$0.013} & 0.049$\pm$0.002 & 0.465$\pm$0.056 & 0.581$\pm$0.031 \\
\checkmark &            & \checkmark & \checkmark & \textbf{0.093$\pm$0.002} & \textbf{0.814$\pm$0.013} & \underline{0.048$\pm$0.001} & \underline{0.483$\pm$0.052} & \textbf{0.593$\pm$0.026} \\
\checkmark & \checkmark &            & \checkmark & 0.099$\pm$0.004 & 0.791$\pm$0.017 & 0.050$\pm$0.002 & 0.464$\pm$0.070 & 0.580$\pm$0.029 \\
\checkmark & \checkmark &            &            & 0.099$\pm$0.003 & 0.793$\pm$0.018 & 0.051$\pm$0.001 & 0.461$\pm$0.044 & 0.581$\pm$0.027 \\
\checkmark &            &            &            & 0.099$\pm$0.002 & 0.797$\pm$0.014 & 0.051$\pm$0.001 & 0.452$\pm$0.055 & 0.573$\pm$0.032 \\
\bottomrule
\end{tabular*}
\end{table}
\renewcommand{\arraystretch}{1.35}

\subsection{National composition map}
\label{sec:res-production}
\label{sec:res-product}

\begin{figure}[pos=htbp]
\centering
\includegraphics[width=0.82\textwidth]{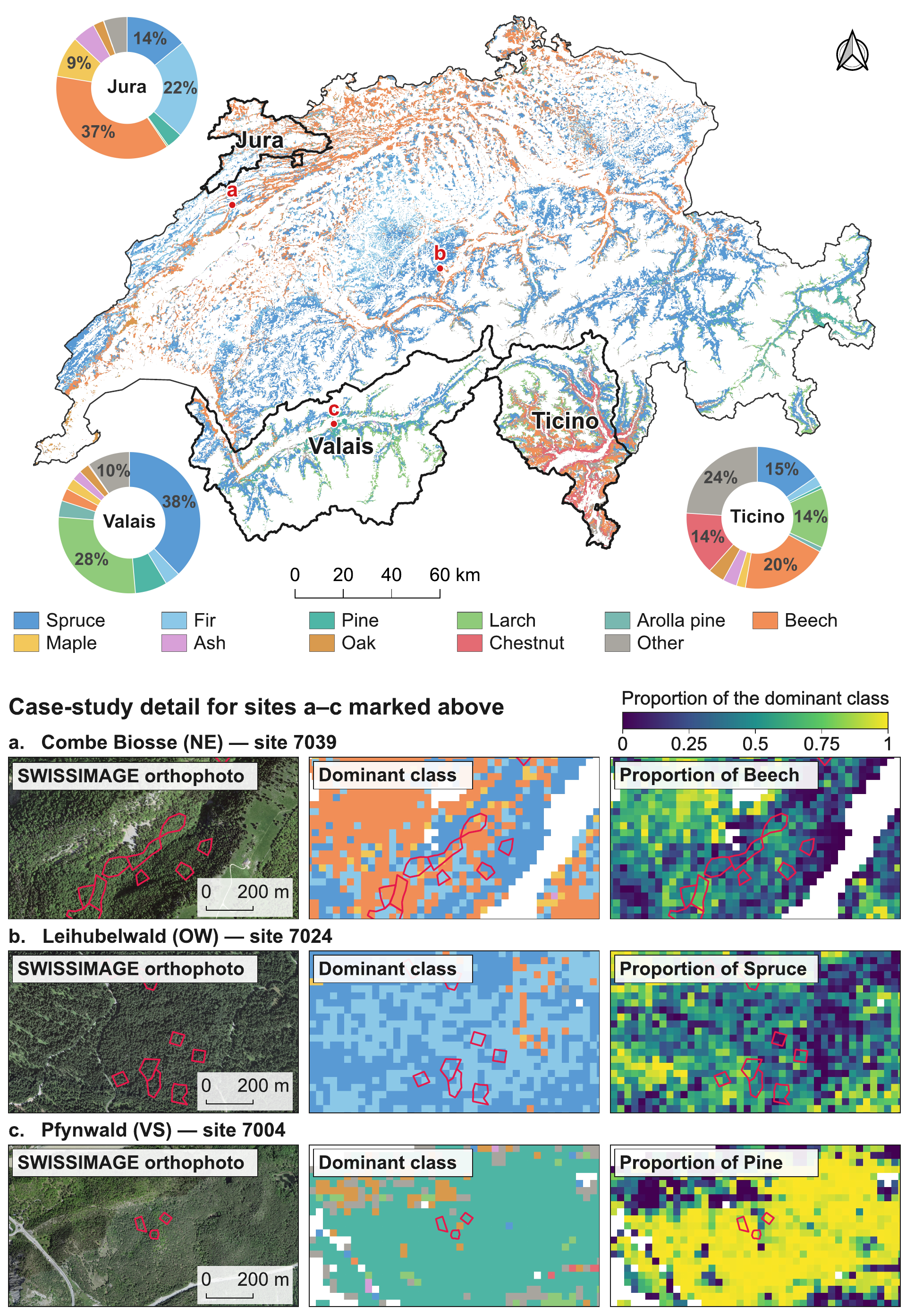}
\caption{Dominant mapped class of the national product. Top: the class with
the largest predicted basal-area share. For display, the 30~m map was
aggregated to 210~m by majority class, and cells with less than 50\%
forest cover are left blank; rings give the composition of three cantons.
Minor classes are shown in Fig.~\ref{fig:national-grid}. Bottom: three forest reserves of
Section~\ref{sec:res-reserves}, marked a--c above, at the native 30~m
resolution. Red outlines are the reserve plot polygons.}
\label{fig:national-argmax}
\end{figure}

\begin{figure}[pos=htbp]
\centering
\includegraphics[width=\textwidth]{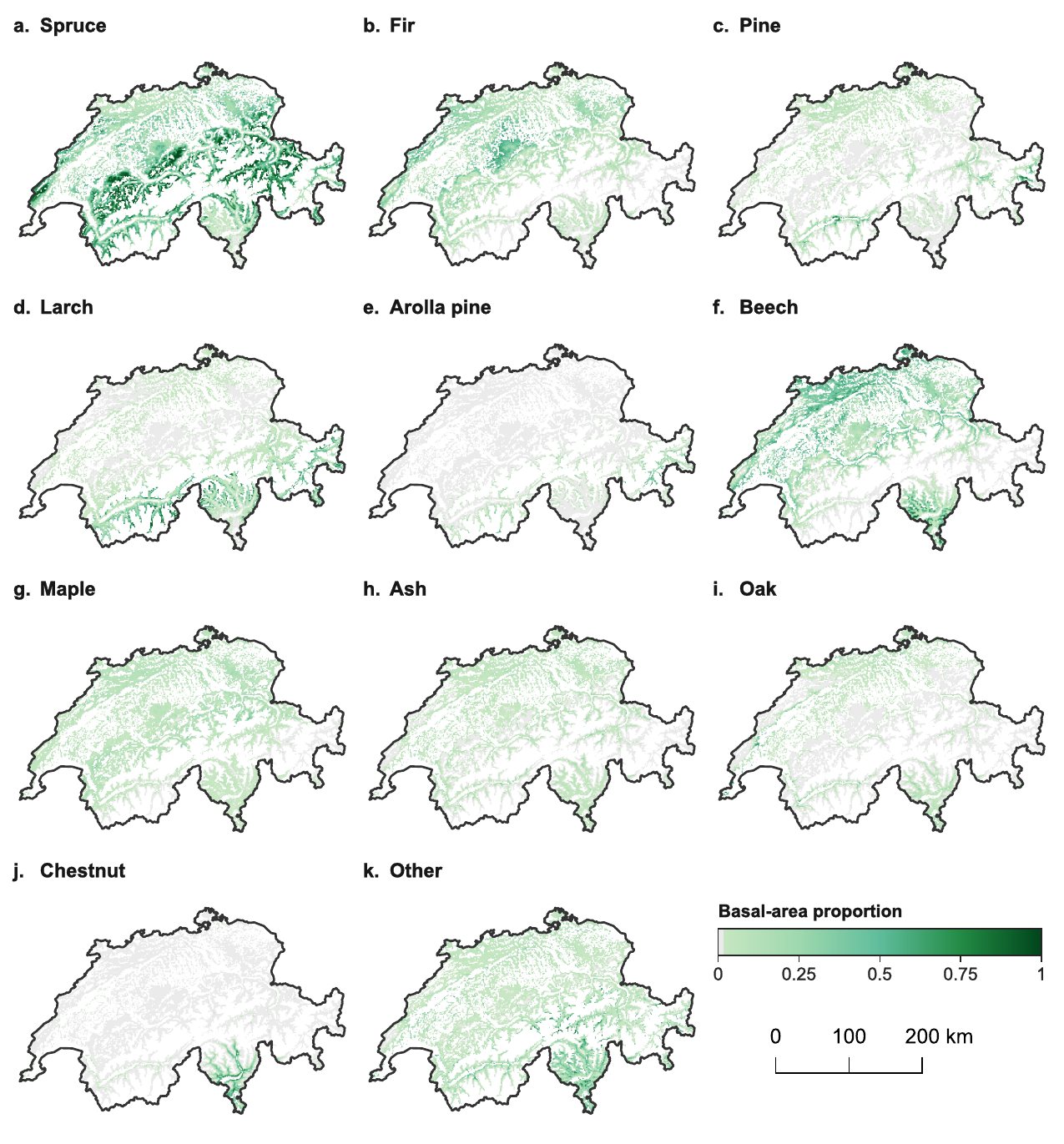}
\caption{Predicted basal-area composition of each mapped class across
Switzerland. Every panel gives the predicted basal-area proportion of one
class on a common linear 0--1 scale: a--e the five coniferous groups, f--j
the five broadleaved groups, and k the residual Other class. For display,
the 30~m predictions were averaged over the forest pixels of 990~m cells,
and only cells with at least 25\% forest cover are shown. Gray marks forest
in which the proportion of the class is below 0.005; areas without forest
are white.}
\label{fig:national-grid}
\end{figure}

Continuous basal-area shares revealed widespread secondary species that
were largely absent from the dominant-class map
(Figs.~\ref{fig:national-argmax} and~\ref{fig:national-grid}). Maple and
ash, for example, rarely formed the dominant class but occurred as minor
components across extensive parts of the Swiss Plateau and Jura. By
contrast, larch, Arolla pine, and chestnut showed more spatially
concentrated distributions and reached high basal-area shares primarily
where they also formed the dominant class. The 30~m product covered
about 13,516~km$^2$ of Swiss forest. At reserve scale, the product
distinguished the fine-grained beech--spruce mosaic at Combe Biosse, the
fir--spruce mixture at Leihubelwald, and the nearly monospecific pine
stand at Pfynwald (Fig.~\ref{fig:national-argmax}).

\subsection{Map evaluation}
\label{sec:res-validation}

\subsubsection{Validation against forest reserve plots}
\label{sec:res-reserves}

Macro-$R^2$ increased from 0.552 at pixel support to 0.733 at plot
support. The pixel-level value was consistent with the spatial
cross-validation result of 0.483 computed over the same ten classes.
Dominant-class accuracy likewise increased from 0.732 at pixel support to
0.811 at plot support and was 0.734 in the spatial cross-validation
(Table~\ref{tab:support-comparison}).

\begin{table}[pos=htbp]
\caption{Accuracy across evaluation supports. $R^2$ is averaged over the
ten classes other than chestnut, and MAE and RMSE over all 11 classes.}
\label{tab:support-comparison}
\small
\renewcommand{\arraystretch}{1.25}
\begin{tabular*}{\textwidth}{@{\extracolsep{\fill}}lccc@{}}
\toprule
\textbf{Metric} & \textbf{Spatial 5-fold CV} & \textbf{Independent, plot-level} & \textbf{Independent, pixel-level} \\
 & NFI ($n=3{,}992$) & Forest reserve ($n=132$) & Forest reserve ($n=414$) \\
\midrule
$R^2$ & 0.483 & 0.733 & 0.552 \\
MAE & 0.048 & 0.038 & 0.042 \\
RMSE & 0.125 & 0.088 & 0.109 \\
Dominant-class accuracy & 0.734 & 0.811 & 0.732 \\
Dominant-class $F_1$ & 0.585 & 0.704 & 0.596 \\
\bottomrule
\end{tabular*}
\end{table}
\renewcommand{\arraystretch}{1.35}

\begin{figure}[pos=htbp]
\centering
\includegraphics[width=\textwidth]{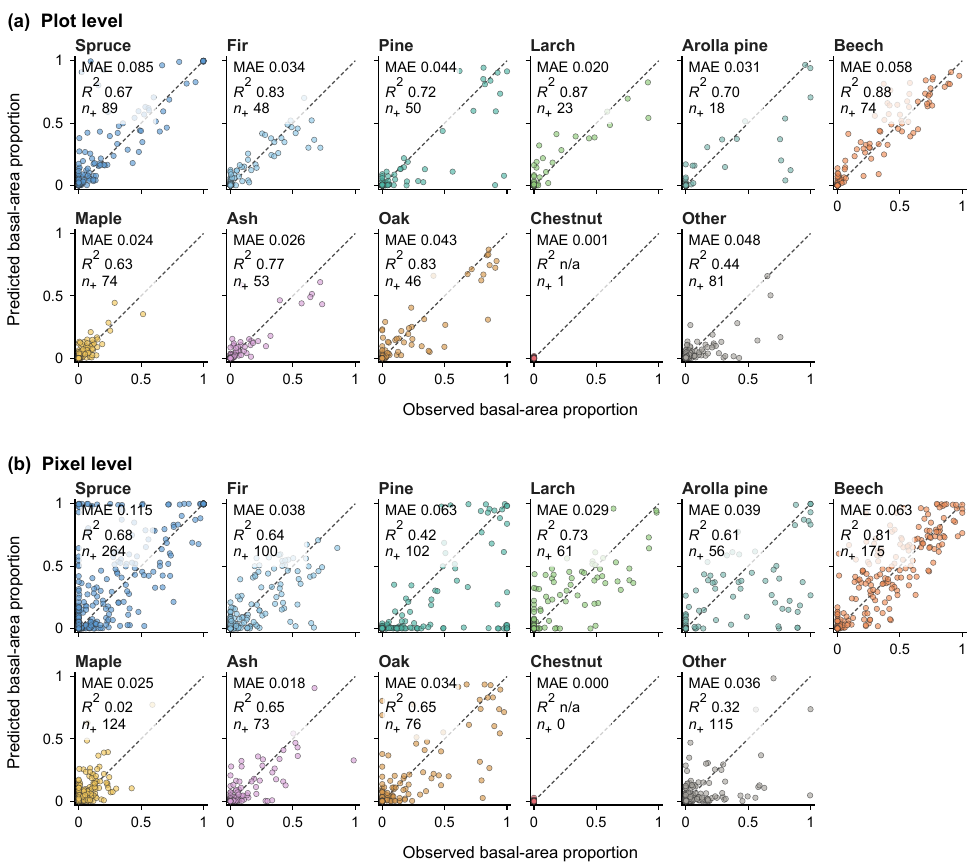}
\caption{Observed and predicted basal-area proportion of each mapped
class. (a) At the 132 forest reserve plots (plot-level support). (b) At
414 forest-reserve pixels, reconstructed on the product's native 30~m
grid (pixel-level support). The dashed line is the 1:1 relationship.
Panel annotations give the per-class MAE, $R^2$, and the number of plots
or pixels in which the class was observed ($n_{+}$). n/a: $R^2$ not evaluable.}
\label{fig:reserve-scatter}
\end{figure}

Composition performance varied across classes and spatial supports
(Fig.~\ref{fig:reserve-scatter}). Beech and larch maintained high $R^2$ at
both plot and pixel support, reaching 0.880 and 0.811 for beech and 0.869
and 0.728 for larch, respectively. Spruce had the largest absolute errors
at both supports (MAE = 0.085 and 0.115). Other had the lowest plot-level
$R^2$ (0.439), whereas maple had the lowest pixel-level $R^2$ (0.021),
despite its low pixel-level MAE of 0.025. Maple and pine showed the largest
drops from plot to pixel support, from 0.633 to 0.021 and from 0.718 to
0.421 in $R^2$.

\begin{figure}[pos=htbp]
\centering
\includegraphics[width=\textwidth]{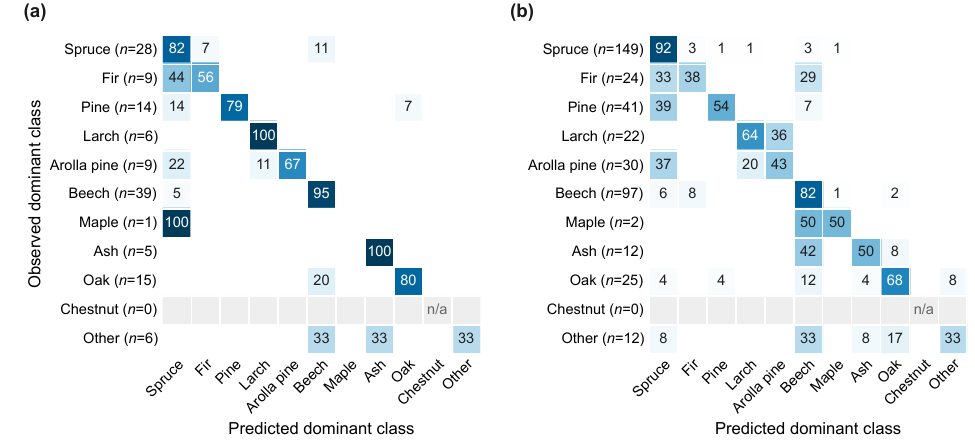}
\caption{Row-normalized confusion matrix of the dominant class at (a)
the 132 forest reserve plots (plot-level support) and (b) the same 414
forest-reserve pixels (pixel-level support). Cell labels give the
percentage of the observed row total; $n$ in the row labels is the
number of plots or pixels in which each class was the observed dominant
one. Chestnut never dominated a plot or pixel and is therefore not
evaluable as an observed class.}
\label{fig:reserve-confusion}
\end{figure}

The producer's accuracy of the dominant class also depended on spatial
support (Fig.~\ref{fig:reserve-confusion}). At plot support, it was highest
for larch and ash (100\% each) and beech (95\%), whereas spruce ranked
highest at pixel support (92\%). It declined from plot to pixel support for
several classes, including pine (79\% to 54\%) and fir (56\% to 38\%).

\subsubsection{Consistency with NFI4 regional statistics}
\label{sec:res-nfi4}

\begin{figure}[pos=htbp]
\centering
\includegraphics[width=\textwidth]{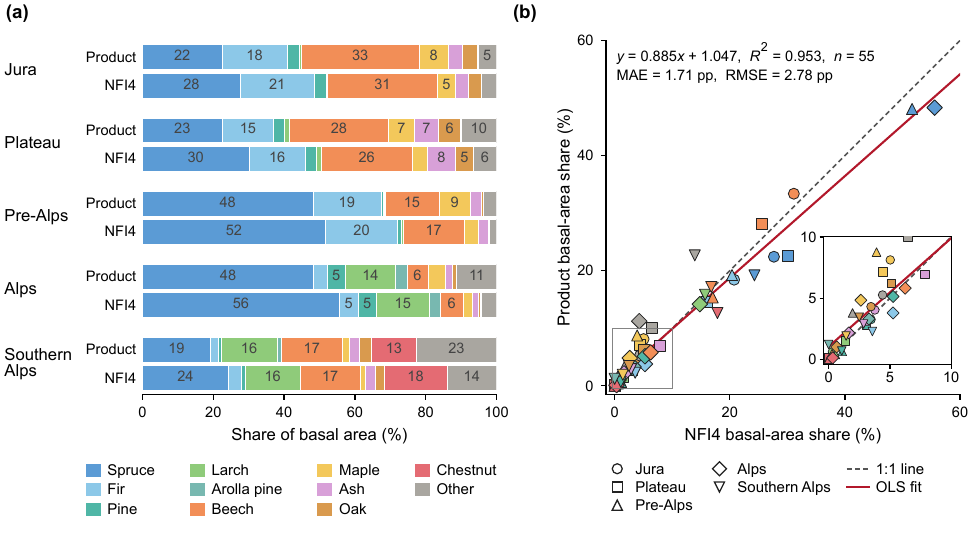}
\caption{Agreement between the national product and NFI4
\citep{abegg2023nfitables} across the five Swiss production regions
(Fig.~\ref{fig:production-regions}).
(a) Basal-area composition of each region in the product and in NFI4;
segments below 5\% are left unlabeled. (b) The same 55 region-by-class
pairs, with color denoting the mapped class and marker shape the production
region. The inset enlarges the range below 10\%, in which most pairs lie;
the dashed line is the 1:1 relationship and the solid line the ordinary
least-squares fit, and MAE and RMSE are taken about the 1:1 line in
percentage points (pp).}
\label{fig:nfi4-regions}
\end{figure}

Regional species shares closely matched the NFI4 statistics
($R^2=0.953$; mean absolute difference = 1.71 percentage points;
Fig.~\ref{fig:nfi4-regions}). Agreement was
highest in the Pre-Alps and lowest in the Southern Alps, with regional
mean absolute differences ranging from 1.27 to 2.18 percentage points.
Across the 55 region-by-class combinations, 70.9\% agreed within 2
percentage points and 87.3\% within 5, and the dominant class agreed in
three of the five production regions.

Spruce showed the largest consistent negative bias across the production
regions ($-3.6$ to $-7.5$ percentage points), whereas Other showed the
largest positive bias ($+0.9$ to $+8.7$ percentage points); maple was also
overestimated by 2.7 percentage points on average. Pine, larch, Arolla
pine, and ash differed least, with mean absolute differences below 0.4
percentage points. The regression slope of 0.885
indicated a moderate compression of the predicted composition range:
shares below 5\% were overestimated by 0.70 percentage points on average,
whereas shares above 15\% were underestimated by 2.52 points.

\subsection{Composition along the elevation gradient}
\label{sec:res-elevation}

\begin{figure}[pos=htbp]
\centering
\includegraphics[width=\textwidth]{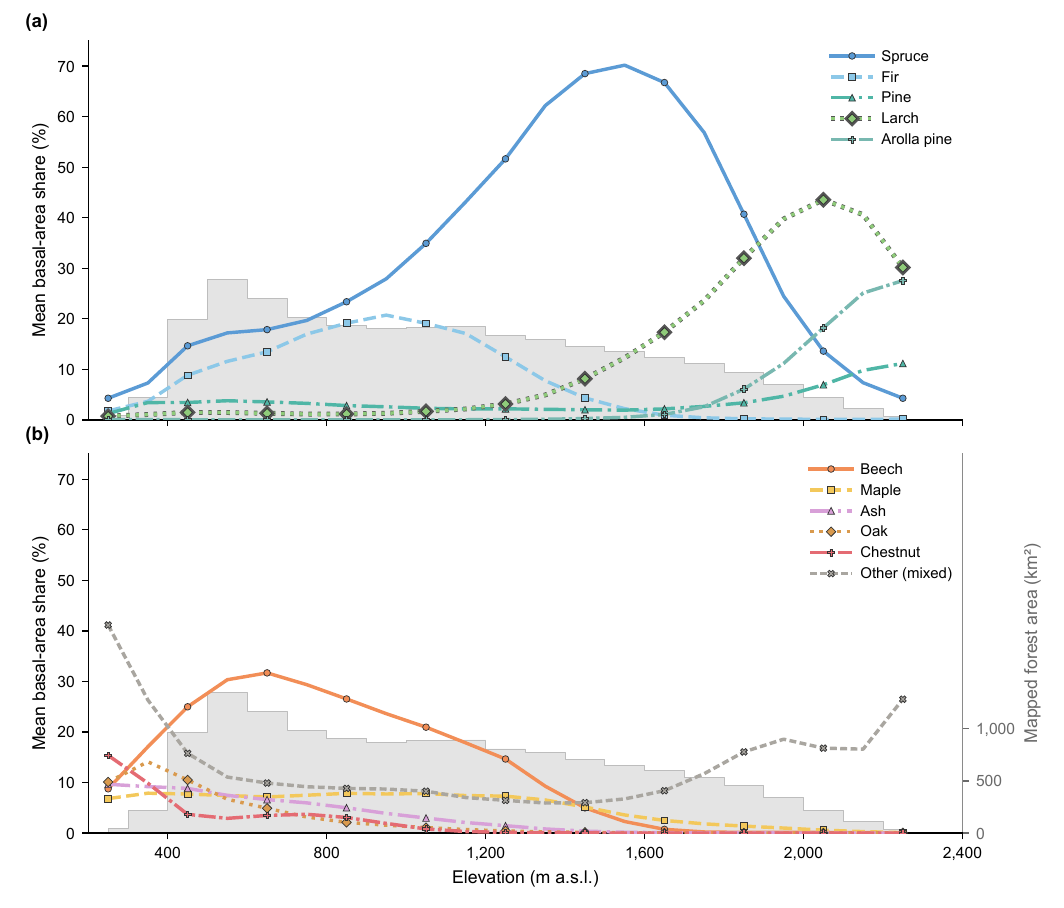}
\caption{Predicted tree-species composition along the Swiss elevation
gradient. (a) The five coniferous groups. (b) The five broadleaved groups
and the residual Other class, sharing the elevation axis with (a). Curves
give the mean predicted basal-area share of each class in each elevation
band, with elevation from swissALTIRegio. The gray backdrop, read against
the right-hand axis in (b), gives the mapped forest area of each band. Only
bands containing at least 5,000 forest pixels (4.5~km$^2$) are shown.}
\label{fig:elevation}
\end{figure}

Dominance shifted from broadleaved species at low elevations to spruce at
middle and high elevations and finally to larch and Arolla pine near the
upper forest limit (Fig.~\ref{fig:elevation}). Beech peaked at 32\% around
650~m, while oak, chestnut, ash, and maple were concentrated in the lowest
bands. Spruce became the largest single component above about 950~m and
reached 70\% around 1,550~m, with fir forming a narrower montane band
beneath it. Larch and Arolla pine together reached about 66\% around
2,150~m. Pine remained a minor component but increased gradually toward
higher elevations, whereas Other was prominent at both ends of the gradient.

\section{Discussion}
\label{sec:discussion}

\subsection{Value of continuous composition maps}
\label{sec:disc-value}

Existing national products for Switzerland describe either the dominant
tree species \citep{koch2024switzerland} or the leaf type
\citep{waser2017walltowall,waser2025foresttype}. Our product adds the
basal-area share of each mapped class, from which a dominant-class map can
still be derived (Fig.~\ref{fig:national-argmax}). This difference matters
most for species that are common but rarely dominant. Maple and ash, for
example, occurred as minor components across much of the Plateau and Jura
but seldom formed the dominant class, so they are largely invisible in a
dominant-class map. These species contribute to the structural and
functional diversity of mixed stands, and their presence can influence how
a stand responds when its dominant species declines.

The product also complements the NFI. The inventory provides design-based
statistics for regions from a sample of plots, whereas the map provides
spatially continuous estimates for every 30~m forest cell. The close
agreement with NFI4 regional shares indicates that the map preserves
aggregate abundance. It can therefore be used where the NFI sample is too
sparse, for example to describe composition within individual forest
districts or along environmental gradients, and as auxiliary information
for model-assisted estimation.

For management, continuous shares allow mixed stands to be characterized
by their actual mixture rather than by a single label. In the montane
belt, spruce, beech, and fir co-occurred in comparable proportions
(Fig.~\ref{fig:elevation}), a situation that a dominant-class map reduces
to whichever species is marginally largest. Such information can help to
identify stands in which a vulnerable species dominates, or in which
admixed species already provide options for climate-adaptive silviculture.

\subsection{Independent generalization and scale-dependent accuracy}
\label{sec:disc-generalization}

The map generalized to the independent forest-reserve inventory, which
entered no stage of model development. At the native 30~m support, its
accuracy was comparable to, and in $R^2$ slightly higher than, that of the
spatial cross-validation. Two differences between the reference datasets
may explain why the independent evaluation was not worse. The reserve
inventories record every tree within a plot polygon, whereas NFI labels are
derived from concentric sample circles in which trees below 36~cm DBH are
recorded on only 200~m$^2$ and weighted by expansion factors, so the NFI
labels carry more sampling noise. The reserve pixels also had to contain at
least 10 trees, which excludes the sparsest cells. Accuracy increased
further at plot support, consistent with the smoothing effect of averaging
predictions over several pixels. Validation results should therefore be
interpreted at the spatial support at which the product is used
\citep{ohmann2014scale,koma2025validation}: plot-level evidence for
aggregated applications, and the pixel-level or cross-validation estimates
for individual pixels.

The comparison with NFI4 complements this site-specific evidence by showing
that the map also preserves regional species abundance. Similar agreement
between remotely sensed species maps and NFI-based regional proportions has
been reported elsewhere, with differences depending on inventory sample
size and the estimator \citep{francini2024forest}. The remaining
disagreement was systematic: spruce was underestimated, Other was
overestimated, and the regression slope of 0.885 indicated moderate
compression toward intermediate shares. Part of this pattern may arise from
the estimands. NFI4 shares are ratios of regional mean basal-area densities
and are therefore weighted toward stands with a high basal area, whereas
map shares are unweighted means of pixel-level proportions.
Spruce-dominated stands often carry a high basal area, so this weighting
would raise the NFI4 share of spruce relative to the map. Heterogeneity
within Other and regression toward the mean may also contribute.

\subsection{Ecologically consistent elevational zonation}
\label{sec:disc-elevation}

The mapped forest area is unevenly distributed across the compositional
zones (Fig.~\ref{fig:elevation}): it peaks between 500 and 700~m, where the
product places beech-dominated stands, and declines steadily above
1,200~m. The elevational sequence the product recovers matches the
classical zonation of Central European mountain forests
\citep{ellenberg2010vegetation}, from oak-, chestnut-, and ash-rich lowland
stands through beech and fir to spruce, and finally to larch and Arolla
pine at the treeline. Between about 800 and 1,200~m, spruce, beech, and fir
each held 17--43\% of the basal area, so this montane belt is where the
mixture information of a continuous product is most informative. The
conifer-dominated belt above roughly 1,200~m coincides with the elevations
at which nationwide standing deadwood peaks \citep{ferrari2026deadwood}, so
a map of where those conifers stand is directly relevant to locating
forests exposed to drought-driven mortality. This agreement should be read
as a check on ecological plausibility rather than as independent evidence
of accuracy: the climate branch of the model carries information that is
itself strongly correlated with elevation.

\subsection{Contributions of the input modalities}
\label{sec:disc-modalities}

The RGB-only model reached $R^2=0.452$, and the three
auxiliary modalities together added 0.033. At 0.1~m, the imagery resolves
crown shape, texture, and local stand arrangement, consistent with
\citet{beloiu2023individual}, who distinguished major tree species in Swiss
forests from aerial RGB imagery. A larger image encoder improved accuracy
more than a different pretraining corpus, which suggests that encoder
capacity limits how much of this information is extracted.

Among the auxiliary inputs, AlphaEarth embeddings and climate variables
contributed most. The embeddings summarize a year of satellite
observations and mainly improved the broadleaved--coniferous distinction
(Section~\ref{sec:res-modalities}), in line with \citet{ball2026gfm}, who
found that foundation-model embeddings improved tree-species
discrimination in temperate mountain forests. Climate likely adds
information by constraining which species can occur at a site, given the
strong climatic gradients across Switzerland. ALS added little in the
present configuration. \citet{murray2024estimating} predicted composition
from ALS alone, and \citet{cao2026m3fnet} improved composition estimates by
fusing lidar with satellite imagery. Here, the ALS encoder was trained from
random initialization on 3,992 sites, and the national point density is
moderate (Section~\ref{sec:disc-limitations}), so this result does not show
that ALS lacks value for composition mapping.

\subsection{Limitations and future research}
\label{sec:disc-limitations}
\label{sec:disc-future}

The main limitation is that the structural information carried by ALS
was not fully exploited. The ALS encoder was trained from random
initialization on only 3,992 sites, and the swissSURFACE3D point clouds,
with a mean density of about 15--20~points~m$^{-2}$, resolve fine vertical
canopy structure only coarsely \citep{michalowska2021review}. The first
nationwide campaign was also flown between 2017 and 2024, so the point
cloud and the aerial image of a site can be several years apart. These
factors may partly explain the confusion among spruce, fir, and Arolla
pine (Fig.~\ref{fig:reserve-confusion}), three dark-foliage conifers that
are difficult to separate by tone and texture in RGB imagery. Pretraining
the ALS encoder on unlabeled point clouds, or retraining it on the second
nationwide survey that swisstopo plans to complete by 2030
\citep{swisstopo2025swisssurface3dsurvey}, may recover more of this
information.

The reference labels also limit the spatial detail of the product. NFI
labels come from concentric sample circles of 200 and 500~m$^2$ that cover
only part of each 30~m cell, and plot size and positioning errors affect
the agreement between field plots and remotely sensed data
\citep{gobakken2009positioning}. Several national products are delivered at
10~m \citep{blickensdorfer2024national,koch2024switzerland,waser2025foresttype,ruetschi2021shrub},
but a 10~m composition map would require reference labels that support
this finer scale.

\section{Conclusions}
\label{sec:conclusions}

We mapped the basal-area shares of ten main tree species and a residual
Other class across Switzerland at 30~m resolution by combining national
forest-inventory observations with multimodal remote sensing and climate
data. Aerial RGB imagery provided most of the predictive information,
with satellite embeddings and climate data providing additional gains.
Independent forest-reserve inventories and regional NFI4 statistics
supported the map's ability to represent composition across spatial
scales, with higher accuracy when predictions were aggregated over plots.
The continuous shares captured both the broad elevational zonation of
Swiss forests and mixtures obscured by dominant-class maps, including
widespread accompanying species and the co-occurrence of spruce, beech,
and fir in the montane belt.

The map shows that much of the compositional variation in Swiss forests
lies below the level of the dominant species. Maple and ash, for example,
rarely dominate a pixel. Yet they occur as minor components across large
parts of the Plateau and Jura. Such admixtures determine which species a
stand can fall back on if its dominant species declines under drought or
other disturbances. The map resolves the share of each main species at
30~m, making these admixtures visible at the national scale. Composition
can thus be located and compared across the country, extending beyond
regional inventory summaries. Repeated acquisitions could also allow
changes in composition to be followed through time.

\section*{CRediT authorship contribution statement}

\textbf{Yixin Zhou:} Conceptualization, Methodology, Software, Validation,
Formal analysis, Investigation, Data curation, Writing -- original draft,
Visualization.
\textbf{Luca Ferrari:} Data curation, Writing -- review \& editing.
\textbf{Verena C. Griess:} Writing -- review \& editing, Supervision,
Funding acquisition.
\textbf{Nataliia Rehush:} Resources, Writing -- review \& editing.
\textbf{Konrad Schindler:} Conceptualization, Methodology, Writing --
review \& editing, Supervision.
\textbf{Mirela Beloiu:} Conceptualization, Methodology, Writing -- review
\& editing, Supervision, Project administration, Funding acquisition.

\section*{Declaration of competing interest}

The authors declare that they have no known competing financial interests
or personal relationships that could have appeared to influence the work
reported in this paper.

\section*{Declaration of generative AI and AI-assisted technologies in the
manuscript preparation process}

During the preparation of this work, the authors used Claude (Anthropic),
an AI-assisted tool, to correct and improve the clarity of sentences, as
well as ChatGPT and Claude Code for programming assistance. After using
these tools, the authors reviewed and edited the content as needed and take
full responsibility for the content of the publication.

\section*{Funding information}

This research was supported by the UPSCALE project, funded by the Swiss
National Science Foundation (SNSF; Sinergia grant CRSII5\_216562).

\section*{Code and data availability}

The national-scale tree species composition maps produced in this study
can be explored interactively through the TreeAI web platform at
\url{https://www.treeai.earth/swiss-species-composition}.
The maps data are openly available on Zenodo at
\url{https://doi.org/10.5281/zenodo.23263490}.
The Swiss National Forest Inventory data used for model training are not
publicly available; they were provided by the Swiss Federal Institute for
Forest, Snow and Landscape Research WSL \citep{wsl2023nfidata} and can be
requested from WSL. SWISSIMAGE aerial orthophotos, swissSURFACE3D point
clouds, and the swissALTIRegio elevation model are freely available from
the Federal Office of Topography swisstopo
(\url{https://www.swisstopo.admin.ch}).
The code used for model training and evaluation is openly available on
GitHub at
\url{https://github.com/yixin-zhou/swiss-tree-species-mapping}.

\section*{Acknowledgements}

Yixin Zhou and Mirela Beloiu warmly thank the Forest Resources Management
group at ETH Zurich for the insightful discussions, as well as Normand
Overney for the support in developing the code to download and process the
data. We also thank Ruoyi Chen for developing the interactive web platform
for visualizing the tree species composition maps. The long-term forest
reserve data of Switzerland were collected in a joint project of the Swiss
Federal Institute for Forest, Snow and Landscape Research WSL and the
Professorship of Forest Ecology of ETH Zurich, with financial support from
the Swiss Federal Office for the Environment. We thank Martina Hobi for providing access to these data.

\bibliographystyle{cas-model2-names}
\bibliography{references}

\clearpage
\appendix
\setcounter{section}{0}\setcounter{figure}{0}\setcounter{table}{0}\setcounter{equation}{0}
\renewcommand{\thesection}{S\arabic{section}}
\renewcommand{\thefigure}{S\arabic{figure}}
\renewcommand{\thetable}{S\arabic{table}}
\renewcommand{\theequation}{S\arabic{equation}}
\renewcommand{\theHsection}{S\arabic{section}}
\renewcommand{\theHfigure}{S\arabic{figure}}
\renewcommand{\theHtable}{S\arabic{table}}
\renewcommand{\theHequation}{S\arabic{equation}}
\begin{center}{\Large\bfseries Supplementary material}\end{center}
\noindent Section, figure, table, and equation numbers prefixed with S refer to this Supplementary material; all other numbers refer to the main text.
\section{NFI field assessment protocol}
\label{sec:appendix-protocol}

Approximately one-ninth of the permanent NFI plots are assessed each year
in annually rotating panels, allowing inventory statistics to be updated
continuously \citep{fischertraub2019nfi}.

At each NFI sample point, field crews record trees within two concentric
circular plots defined by DBH (measured 1.3~m above ground): all trees and shrubs with DBH $\geq$ 12~cm are recorded
within an inner 200~m$^2$ circle (7.98~m radius), and all trees and shrubs
with DBH $\geq$ 36~cm are additionally recorded within an outer 500~m$^2$
circle (12.62~m radius)
\citep{fischertraub2019nfi,duggelin2020nfi5}. Species identity, DBH, canopy position, and
coordinates are recorded for every measured tree; canopy-layer assignment
follows relative height within the stand, stratified into an upper layer
(Oberschicht, $\geq$2/3), a middle layer (Mittelschicht, 1/3--2/3), and an
understory (Unterschicht, $\leq$1/3). The upper and middle strata are retained for the labels in
Section~\ref{sec:labels}.
A detailed description of the NFI sampling design is available in the
official inventory-concept documentation
\citep{swissnfi2026inventoryconcept}.

The nested design gives different spatial support to different stem-size
classes: stems with DBH 12--35.9~cm are sampled within 200~m$^2$, whereas
stems with DBH $\geq$36~cm are observed over 500~m$^2$. The 28~m
RS crop slightly exceeds the 25.2~m diameter of the outer circle
and therefore contains the complete nominal field footprint while allowing
for coordinate uncertainty and edge crowns. Composition targets account for
the unequal sampling support by multiplying individual-tree basal area by the
corresponding per-hectare expansion factor (50 or 20) before aggregation.
They should therefore be interpreted as expanded basal-area composition
rather than as an unweighted census of stems across 500~m$^2$.

For tree $j$ at site $i$, unexpanded basal area is calculated from DBH in
centimeters as

\begin{equation}
b_{ij}=\pi\left(\frac{\mathrm{DBH}_{ij}}{200}\right)^2,
\label{eq:tree-ba}
\end{equation}

in m$^2$ tree$^{-1}$. Multiplication by the expansion factor gives
$a_{ij}=w_{ij}b_{ij}$ in m$^2$ ha$^{-1}$, as used in
Eq.~\ref{eq:species}.

\section{NFI eligibility filtering}
\label{sec:appendix-attrition}

Sites with stand-replacing mortality or clearing between the NFI survey
and RS acquisitions were excluded. SWISSIMAGE imagery at the remaining
sites was inspected visually, and sites with degraded imagery, most often
due to terrain or cloud shadow, were removed. All retained sites were
required to have complete multimodal records.

Table~\ref{tab:eligibility-attrition} reports the number of tree-level
observations and sites remaining after temporal screening, exclusion of the
lowest canopy stratum, and disturbance and completeness screening. Counts
are cumulative, with each row reflecting all rules applied up to that step.

\begin{table}[pos=htbp]
\caption{Attrition of tree-level observations and sites through the
eligibility filtering pipeline.}
\label{tab:eligibility-attrition}
\setlength{\tabcolsep}{3pt}
\renewcommand{\arraystretch}{1.15}
\renewcommand{\tabularxcolumn}[1]{>{\raggedright\arraybackslash}m{#1}}
\small
\begin{tabularx}{\textwidth}{@{}>{\raggedright\arraybackslash}m{0.8cm}
                              >{\raggedright\arraybackslash}m{3.9cm}
                              X
                              >{\raggedleft\arraybackslash}m{1.5cm}
                              >{\raggedleft\arraybackslash}m{1.1cm}}
\toprule
\textbf{Step} & \textbf{Screening criterion} & \textbf{Description} & \textbf{Tree points} & \textbf{Sites} \\
\midrule
1 & Raw NFI tree observations & All available tree records & 63,969 & 5,998 \\
2 & Recent inventory observations & Recorded in 2017 or later & 42,831 & 4,052 \\
3 & Canopy-layer screening & Exclude the lowest stratum & 39,863 & 4,043 \\
4 & Quality control and completeness & Exclude disturbed sites and sites with degraded imagery; require complete multimodal records & 39,380 & 3,992 \\
\bottomrule
\end{tabularx}
\end{table}
\renewcommand{\tabularxcolumn}[1]{p{#1}}
\renewcommand{\arraystretch}{1.35}

The full sequence retained 66.6\% of the original sites and 61.6\% of the
tree records. Temporal screening caused the largest reduction, removing
1,946 sites. Excluding the lowest stratum removed only nine additional sites
but 2,968 tree records, because excluded stems commonly co-occurred with
retained upper- and middle-stratum trees. Disturbance, image-quality, and
completeness screening removed a further 51 sites and 483 tree records.

\section{Examples of reference target construction}
\label{sec:appendix-labels}

\begin{figure}[pos=htbp]
\centering
\includegraphics[width=\textwidth]{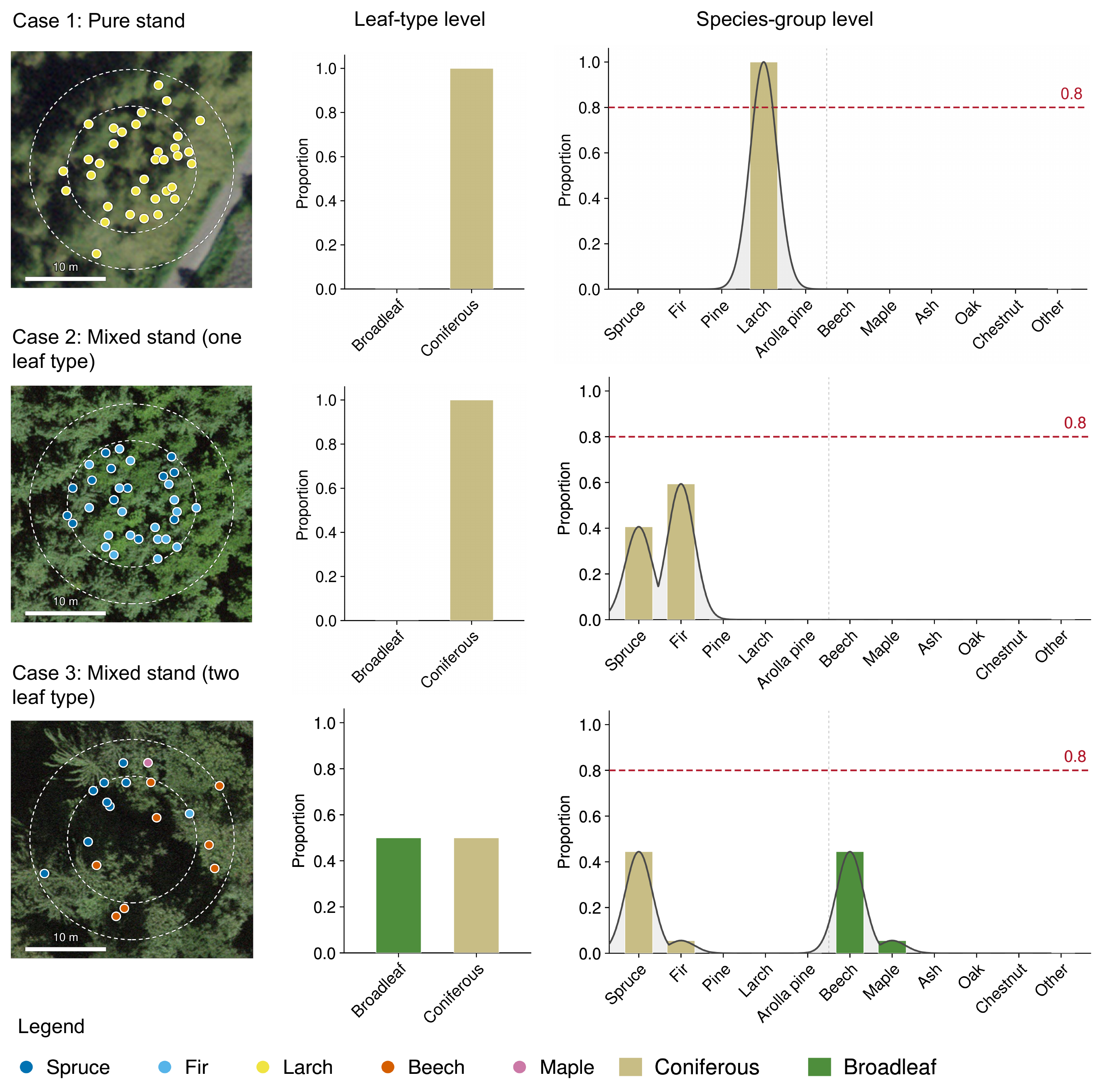}
\caption{Conceptual examples of the three principal site-level basal-area
distribution forms and their complementary targets. The examples represent a
pure stand, a mixed stand within one leaf type, and a mixed stand spanning
both leaf types. Left: aerial imagery and NFI sample circles; center:
schematic class composition and the 0.80 stand-type threshold;
right: the corresponding broadleaved--coniferous composition.}
\label{fig:label-construction}
\end{figure}

Figure~\ref{fig:label-construction} illustrates the species-composition
label of Section~\ref{sec:labels} and the two auxiliary targets of
Section~\ref{sec:model-heads}. The examples show dominance by one mapped
class, mixtures within one leaf type, and mixtures spanning both leaf types.

\section{GPS positional extent of NFI tree records}
\label{sec:appendix-gps-extent}

National tree-species products for neighboring countries are increasingly
delivered at 10~m grid spacing \citep{blickensdorfer2024national}. We
characterized the spatial extent represented by the retained tree positions
at the 3,992 eligible sites (Table~\ref{tab:eligibility-attrition}). Each
site retained a median of 9 trees (interquartile range 5--13; range 1--46).
Although each site has a reference coordinate, every retained tree carries
its own field-recorded position rather than sharing that coordinate.

\begin{figure}[pos=htbp]
\centering
\includegraphics[width=\textwidth]{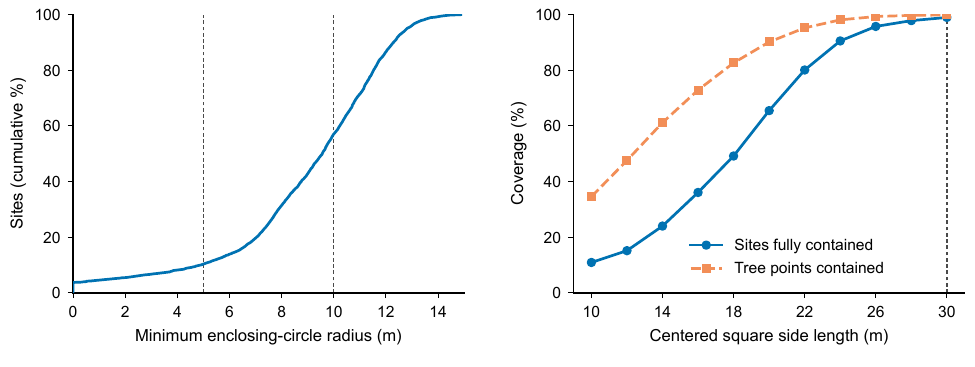}
\caption{GPS positional extent of the retained NFI tree records at 3,992
eligible sites. Left: cumulative distribution of the minimum
enclosing-circle radius of retained tree positions; vertical dashed lines
mark 5~m and 10~m. Right: containment within centered square pixels of
increasing side length; the solid curve gives sites fully contained, the
dashed curve individual tree points contained, and the vertical dashed line
marks the 30~m national-product grid.}
\label{fig:gps-extent}
\end{figure}

Figure~\ref{fig:gps-extent} summarizes this extent in two complementary
ways. The left panel gives the cumulative distribution of the minimum
enclosing-circle radius, with dashed reference lines at 5~m and 10~m. The
radius exceeds 5~m at 89.7\% of sites and 10~m at 42.9\% of sites. The right
panel and Table~\ref{tab:gps-extent-summary} report containment within
centered square grid cells. All retained tree positions are contained within
a 10~m cell at 10.9\% of sites, compared with 65.5\% at 20~m and 99.0\% at
30~m. The corresponding proportions of individual tree positions are 34.6\%,
90.2\%, and 99.9\%, respectively.

\begin{table}[pos=htbp]
\caption{Containment of retained NFI tree positions within a square output
pixel of the given side length, centered on the site coordinate, for the
3,992 eligible sites.}
\label{tab:gps-extent-summary}
\setlength{\tabcolsep}{3pt}
\renewcommand{\arraystretch}{1.15}
\small
\begin{tabular*}{\textwidth}{@{\extracolsep{\fill}}lccccccc@{}}
\toprule
\textbf{Pixel side length} & \textbf{10~m} & \textbf{14~m} & \textbf{18~m} & \textbf{20~m} & \textbf{24~m} & \textbf{30~m (used here)} \\
\midrule
Sites fully contained (\%) & 10.9 & 24.0 & 49.2 & 65.5 & 90.6 & 99.0 \\
Tree points contained (\%) & 34.6 & 61.2 & 82.7 & 90.2 & 98.1 & 99.9 \\
\bottomrule
\end{tabular*}
\end{table}

These summaries characterize the spatial support of the field-recorded tree
positions, rather than independently validating a particular map resolution.
They provide context for the 30~m grid used in this study and for future
products at finer spatial resolution.

\section{Reference year of the national inference}
\label{sec:appendix-inference-year}

\begin{figure}[pos=htbp]
\centering
\includegraphics[width=\textwidth]{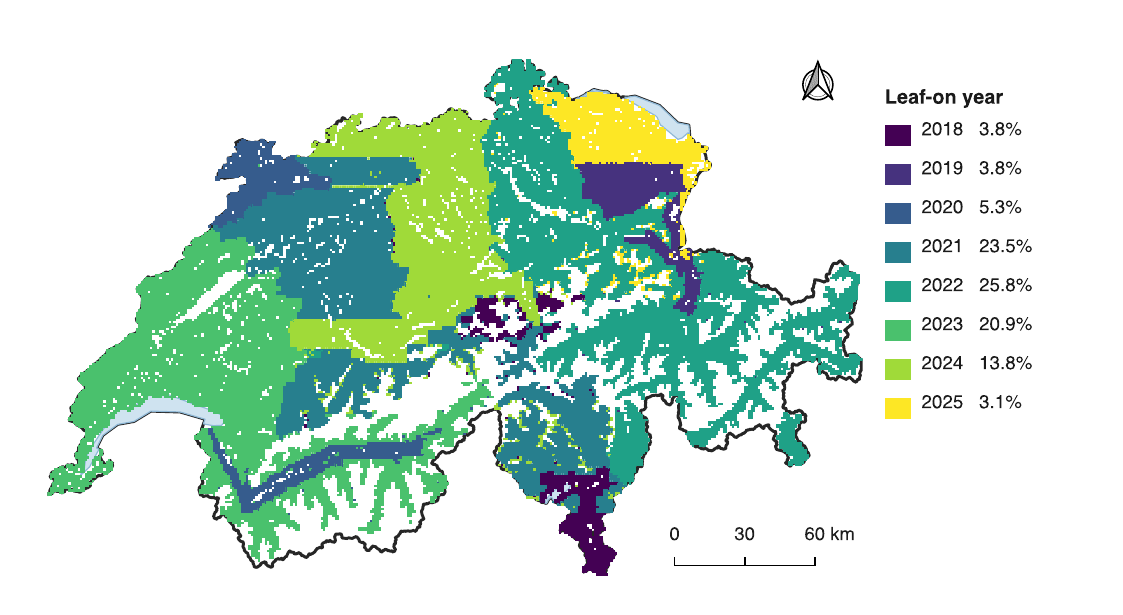}
\caption{Reference year of the national-scale inference. Each forest pixel was
assigned the year of the most recent leaf-on SWISSIMAGE acquisition
available for its area, and that year also determined which annual
AlphaEarth embedding was used. Percentages give the share of the mapped
forest area falling in each year.}
\label{fig:inference-year}
\end{figure}

SWISSIMAGE is renewed region by region, so the national inference data
described in Section~\ref{sec:inference-data} do not share a single
acquisition date. Fig.~\ref{fig:inference-year} maps the reference year
assigned to each forest pixel and the share of the mapped forest area
falling in each year.
Across the mapped forest area, 2021--2024 account for 84.0\% of the
area, while 2018 and 2025 account for 3.8\% and 3.1\%, respectively.

\section{Definition of the mapped classes}
\label{sec:appendix-classes}

Table~\ref{tab:species-groups} lists the 11 mapped classes and the NFI
taxa included in each.

\begin{table}[pos=htbp]
\caption{Definition of the 11 mapped classes. The ten main tree species
follow the Swiss NFI grouping; Other contains all retained taxa outside
those ten. English and Latin give the common and scientific name of each
class, respectively.}
\label{tab:species-groups}
\setlength{\tabcolsep}{3pt}
\renewcommand{\arraystretch}{1.15}
\renewcommand{\tabularxcolumn}[1]{>{\raggedright\arraybackslash}m{#1}}
\normalsize
\begin{tabularx}{\textwidth}{@{}>{\raggedright\arraybackslash}m{0.75cm}
                              @{\hspace{8pt}}>{\raggedright\arraybackslash}m{2.2cm}
                              >{\raggedright\arraybackslash}m{2.6cm}
                              X}
\toprule
\textbf{No.} & \textbf{English} & \textbf{Latin} & \textbf{Included species} \\
\midrule
1 & Spruce & \textit{Picea} spp. & \textit{Picea abies}; \textit{Picea spec.} \\
2 & Fir & \textit{Abies alba} & \textit{Abies alba} \\
3 & Pine & \textit{Pinus} spp. & \textit{Pinus sylvestris}; \textit{Pinus nigra}; \textit{Pinus strobus}; \textit{Pinus mugo arborea} \\
4 & Larch & \textit{Larix} spp. & \textit{Larix decidua}; \textit{Larix kaempferi} \\
5 & Arolla pine & \textit{Pinus cembra} & \textit{Pinus cembra} \\
6 & Beech & \textit{Fagus sylvatica} & \textit{Fagus sylvatica} \\
7 & Maple & \textit{Acer} spp. & \textit{Acer campestre}; \textit{Acer platanoides}; \textit{Acer pseudoplatanus}; \textit{Acer opalus} \\
8 & Ash & \textit{Fraxinus} spp. & \textit{Fraxinus excelsior}; \textit{Fraxinus ornus} \\
9 & Oak & \textit{Quercus} spp. & \textit{Quercus robur}; \textit{Quercus petraea}; \textit{Quercus pubescens}; \textit{Quercus cerris}; \textit{Quercus rubra} \\
10 & Chestnut & \textit{Castanea sativa} & \textit{Castanea sativa} \\
11 & Other & --- & --- (48 mixed broadleaf and coniferous taxa, not itemized) \\
\bottomrule
\end{tabularx}
\end{table}
\renewcommand{\tabularxcolumn}[1]{p{#1}}
\renewcommand{\arraystretch}{1.35}

\section{Species-to-leaf-type mapping}
\label{sec:appendix-leaftype}

Each of the 71 NFI taxa retained after site and canopy-layer screening was
assigned to either broadleaved or coniferous
(Table~\ref{tab:leaf-type-mapping}). The assignment was made at the original
taxon level, before the taxa were aggregated into the 11 mapped classes. The
mapping contains 57 broadleaved and 14 coniferous taxa. Importantly,
the residual Other class contains 48 taxa, including 44 broadleaved and four
coniferous taxa; leaf type therefore cannot be inferred from the model-class
label alone.

\begin{table}[pos=htbp]
\caption{Assignment of the 71 retained NFI taxa to broadleaved or
coniferous leaf type.}
\label{tab:leaf-type-mapping}
\setlength{\tabcolsep}{3pt}
\renewcommand{\arraystretch}{1.1}
\small
\begin{tabularx}{\textwidth}{@{}>{\raggedright\arraybackslash}X
                              >{\raggedright\arraybackslash}p{2.2cm}
                              >{\raggedright\arraybackslash}X
                              >{\raggedright\arraybackslash}p{2.2cm}}
\toprule
\textbf{Taxon} & \textbf{Leaf type} & \textbf{Taxon} & \textbf{Leaf type} \\
\midrule
\textit{Abies alba} & Coniferous & \textit{Pinus sylvestris} & Coniferous \\
\textit{Acer campestre} & Broadleaved & \textit{Platanus spec.} & Broadleaved \\
\textit{Acer opalus} & Broadleaved & \textit{Populus alba} & Broadleaved \\
\textit{Acer platanoides} & Broadleaved & \textit{Populus nigra s.l.} & Broadleaved \\
\textit{Acer pseudoplatanus} & Broadleaved & \textit{Populus spec.} & Broadleaved \\
\textit{Aesculus hippocastanum} & Broadleaved & \textit{Populus tremula} & Broadleaved \\
\textit{Ailanthus altissima} & Broadleaved & \textit{Populus x canescens} & Broadleaved \\
\textit{Alnus glutinosa} & Broadleaved & \textit{Prunus avium} & Broadleaved \\
\textit{Alnus incana} & Broadleaved & \textit{Prunus domestica} & Broadleaved \\
\textit{Alnus viridis} & Broadleaved & \textit{Prunus padus} & Broadleaved \\
\textit{Betula pendula} & Broadleaved & \textit{Prunus serotina} & Broadleaved \\
\textit{Betula pubescens} & Broadleaved & \textit{Pseudotsuga menziesii} & Coniferous \\
\textit{Carpinus betulus} & Broadleaved & \textit{Pyrus communis} / \textit{P. pyraster} & Broadleaved \\
\textit{Castanea sativa} & Broadleaved & \textit{Quercus cerris} & Broadleaved \\
\textit{Celtis australis} & Broadleaved & \textit{Quercus petraea} & Broadleaved \\
\textit{Corylus avellana} & Broadleaved & \textit{Quercus pubescens} & Broadleaved \\
\textit{Crataegus monogyna} & Broadleaved & \textit{Quercus robur} & Broadleaved \\
\textit{Fagus sylvatica} & Broadleaved & \textit{Quercus rubra} & Broadleaved \\
\textit{Fraxinus excelsior} & Broadleaved & \textit{Robinia pseudoacacia} & Broadleaved \\
\textit{Fraxinus ornus} & Broadleaved & \textit{Salix alba} & Broadleaved \\
\textit{Ilex aquifolium} & Broadleaved & \textit{Salix caprea} & Broadleaved \\
\textit{Juglans regia} & Broadleaved & \textit{Salix spec.} & Broadleaved \\
\textit{Laburnum alpinum} & Broadleaved & \textit{Sambucus nigra} & Broadleaved \\
\textit{Laburnum anagyroides} & Broadleaved & \textit{Sorbus aria} & Broadleaved \\
\textit{Larix decidua} & Coniferous & \textit{Sorbus aucuparia} & Broadleaved \\
\textit{Larix kaempferi} & Coniferous & \textit{Sorbus domestica} & Broadleaved \\
\textit{Malus pumila} & Broadleaved & \textit{Sorbus latifolia s.l.} & Broadleaved \\
\textit{Malus sylvestris} & Broadleaved & \textit{Sorbus mougeotii} & Broadleaved \\
\textit{Ostrya carpinifolia} & Broadleaved & \textit{Sorbus torminalis} & Broadleaved \\
\textit{Picea abies} & Coniferous & \textit{Taxus baccata} & Coniferous \\
\textit{Picea spec.} & Coniferous & \textit{Thuja spec.} & Coniferous \\
\textit{Pinus cembra} & Coniferous & \textit{Tilia cordata} & Broadleaved \\
\textit{Pinus mugo arborea} & Coniferous & \textit{Tilia platyphyllos} & Broadleaved \\
\textit{Pinus mugo prostrata} & Coniferous & \textit{Ulmus glabra} & Broadleaved \\
\textit{Pinus nigra} & Coniferous & \textit{Ulmus minor} & Broadleaved \\
\textit{Pinus strobus} & Coniferous & & \\
\bottomrule
\end{tabularx}
\end{table}
\renewcommand{\arraystretch}{1.35}

Leaf-type proportions were calculated by summing the expanded basal area of
all original taxa assigned to each leaf type. Because this aggregation
precedes formation of the 11 mapped classes, the broadleaved--coniferous balance
remains fully defined even though Other combines taxa from both leaf types.
The leaf-type target therefore provides a functional summary of the site
composition independently of the species-composition target.

\section{Sentinel-2 spectral phenology}
\label{sec:appendix-s2phenology}

\begin{figure}[pos=htbp]
\centering
\includegraphics[width=\textwidth]{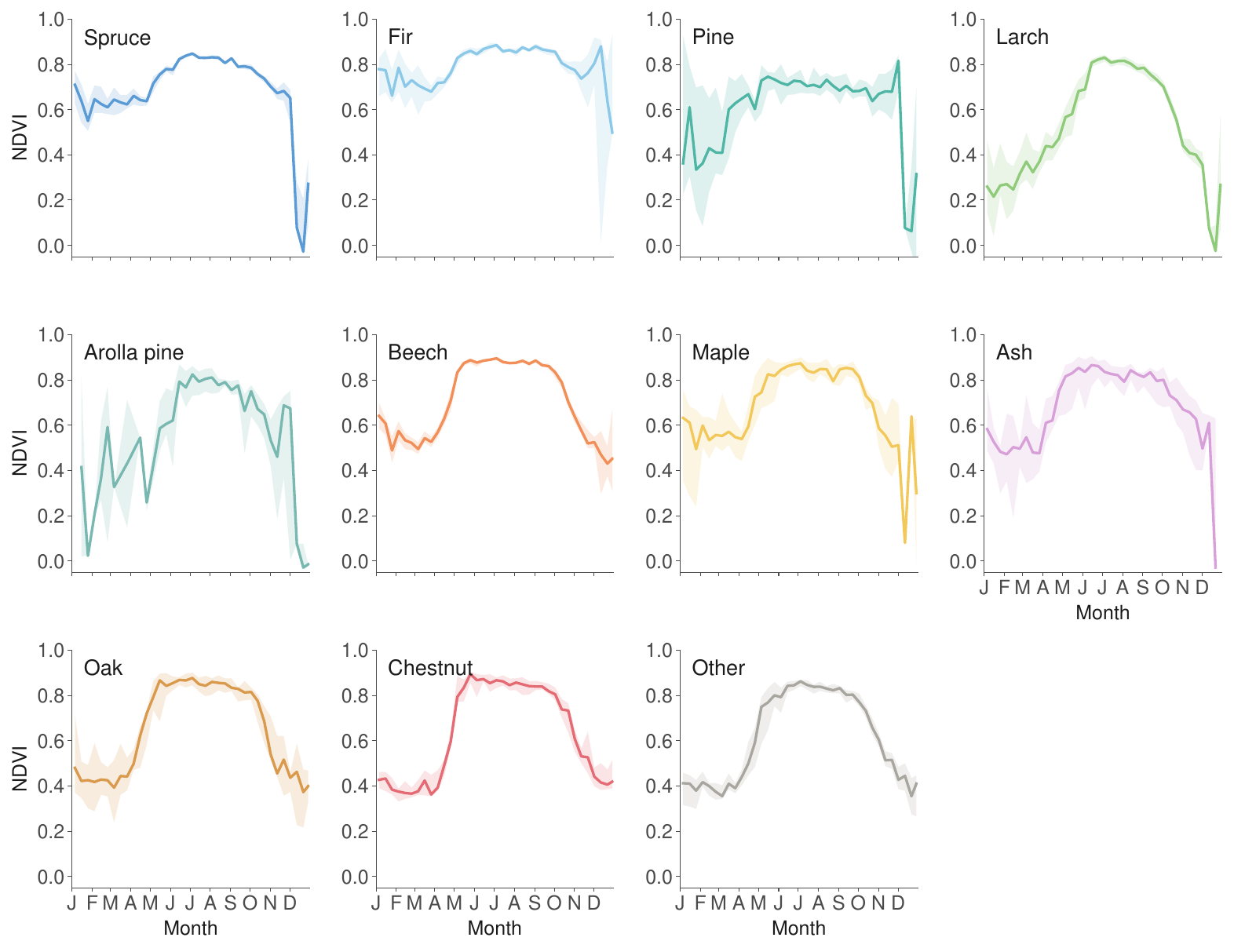}
\caption{Seasonal NDVI trajectories by mapped class, derived from the
quality-screened Sentinel-2 time series. Lines show the class median across
sites; shaded bands show the bootstrap confidence interval of the median.}
\label{fig:ndvi-phenology}
\end{figure}

Fig.~\ref{fig:ndvi-phenology} shows the seasonal NDVI trajectory of each
mapped class, computed from the quality-screened Sentinel-2 time series
described in Section~\ref{sec:appendix-design}.

The trajectories show a comparatively stable evergreen-conifer signal and a
much stronger seasonal amplitude for deciduous canopies. Larch, the only
deciduous conifer among the main tree species, separates from the evergreen
conifers most clearly during leaf emergence and senescence, whereas the
broadleaved groups share a broadly similar summer plateau. These are
descriptive summaries: they indicate which temporal signals are available to
the model, but they do not by themselves establish species-level
separability, because geography, elevation, illumination, and sample size
also vary among groups.

\section{CHELSA climate variables}
\label{sec:appendix-chelsa}

Before entering the climate
encoder, each variable was standardized as $z_j=(x_j-\mu_j)/\sigma_j$ using
its variable-wise mean and standard deviation. The variables and their units
are listed in Table~\ref{tab:chelsa-standardization}.

\begin{table}[pos=htbp]
\centering
\caption{Definitions of the 19 CHELSA BIOCLIM+ variables used by the model.}
\label{tab:chelsa-standardization}
\setlength{\tabcolsep}{3pt}
\renewcommand{\arraystretch}{1.15}
\renewcommand{\tabularxcolumn}[1]{>{\raggedright\arraybackslash}m{#1}}
\small
\begin{tabularx}{0.78\textwidth}{@{}>{\raggedright\arraybackslash}m{1.4cm}
                              X
                              >{\raggedright\arraybackslash}m{2.5cm}}
\toprule
\textbf{Variable} & \textbf{Interpretation} & \textbf{Unit} \\
\midrule
bio01 & Annual mean temperature & $^\circ$C \\
bio02 & Mean diurnal temperature range & $^\circ$C \\
bio03 & Isothermality (bio02/bio07 $\times$ 100) & \% \\
bio04 & Temperature seasonality (standard deviation $\times$ 100) & $^\circ$C $\times$ 100 \\
bio05 & Maximum temperature of the warmest month & $^\circ$C \\
bio06 & Minimum temperature of the coldest month & $^\circ$C \\
bio07 & Annual temperature range (bio05 $-$ bio06) & $^\circ$C \\
bio08 & Mean temperature of the wettest quarter & $^\circ$C \\
bio09 & Mean temperature of the driest quarter & $^\circ$C \\
bio10 & Mean temperature of the warmest quarter & $^\circ$C \\
bio11 & Mean temperature of the coldest quarter & $^\circ$C \\
bio12 & Annual precipitation & mm \\
bio13 & Precipitation of the wettest month & mm \\
bio14 & Precipitation of the driest month & mm \\
bio15 & Precipitation seasonality (coefficient of variation) & \% \\
bio16 & Precipitation of the wettest quarter & mm \\
bio17 & Precipitation of the driest quarter & mm \\
bio18 & Precipitation of the warmest quarter & mm \\
bio19 & Precipitation of the coldest quarter & mm \\
\bottomrule
\end{tabularx}
\end{table}
\renewcommand{\tabularxcolumn}[1]{p{#1}}
\renewcommand{\arraystretch}{1.35}

\section{Spatial fold assignment}
\label{sec:appendix-splits}

The five folds of the spatial cross-validation
(Section~\ref{sec:splits}) contained 799, 798, 796, 800, and 799 sites,
respectively, and had comparable mean compositions
(Fig.~\ref{fig:cv-folds-map}a).

\section{Training implementation details}
\label{sec:appendix-training}

The inputs and encoders of the four modalities described in
Section~\ref{sec:model-encoders} are summarized in Table~\ref{tab:encoders}.

The aerial encoder reads out the class and patch tokens of the blocks
$\mathcal{L}=\{7,15,23,31\}$ of the frozen DINOv3 ViT-H+/16. The tokens
of each block are projected to the common dimension $D$ by the same
linear layer, and the two token types are combined across layers using
separate learnable weights,

\begin{equation}
\begin{aligned}
(\mathbf{t}^{(l)}_{\mathrm{cls}},\mathbf{T}^{(l)}_{\mathrm{patch}})
&=\operatorname{Split}\!\left(\mathbf{W}\,\mathrm{LN}\!\left(\mathbf{Z}^{(l)}\right)\right),\\
\mathbf{t}_{\mathrm{cls}}
&=\sum_{l\in\mathcal{L}}\alpha^{\mathrm{cls}}_l\mathbf{t}^{(l)}_{\mathrm{cls}},
&\mathbf{T}_{\mathrm{patch}}
&=\sum_{l\in\mathcal{L}}\alpha^{\mathrm{patch}}_l\mathbf{T}^{(l)}_{\mathrm{patch}},\\
\boldsymbol{\alpha}^{\mathrm{cls}}
&=\mathrm{softmax}(\mathbf{a}_{\mathrm{cls}}),
&\boldsymbol{\alpha}^{\mathrm{patch}}
&=\mathrm{softmax}(\mathbf{a}_{\mathrm{patch}}).
\end{aligned}
\label{eq:dinov3-combine}
\end{equation}

where $\mathbf{a}_{\mathrm{cls}}$ and $\mathbf{a}_{\mathrm{patch}}$ are
independent learnable four-dimensional vectors. Both are initialized to
zero, giving an initial weight of 0.25 to each block after the softmax. The
patch weights are shared across all patch positions, while the projection
$\mathbf{W}$ is shared by class and patch tokens.

\begin{table}[pos=htbp]
\caption{Modality encoders. GSD = ground sampling distance; MLP = multilayer perceptron.}
\label{tab:encoders}
\setlength{\tabcolsep}{3pt}
\renewcommand{\arraystretch}{1.15}
\renewcommand{\tabularxcolumn}[1]{>{\raggedright\arraybackslash}m{#1}}
\small
\begin{tabularx}{\textwidth}{@{}>{\raggedright\arraybackslash}m{2.4cm}
                              >{\raggedright\arraybackslash}m{3.7cm}
                              X
                              >{\raggedright\arraybackslash}m{1.6cm}}
\toprule
\textbf{Modality} & \textbf{Model input} & \textbf{Encoder} & \textbf{Trainable} \\
\midrule
Aerial imagery & 280$\times$280$\times$3 crop, resized to 256$\times$256 (28~m, 0.1~m GSD) & DINOv3 ViT-H+/16, blocks \{7, 15, 23, 31\}, 8$\times$8 patch pooling & No \\
ALS & Points within 14~m radius, 0.1~m voxels & Sparse 3D U-Net (channel widths 32--160) + 64 learned queries & Yes \\
Satellite embedding & 3$\times$3$\times$64 AlphaEarth patch (10~m pixels, 30~m footprint) & Shared 2-layer MLP + positional embedding & Yes \\
Climate & 19 CHELSA BIOCLIM+ variables & 2-layer MLP & Yes \\
\bottomrule
\end{tabularx}
\end{table}
\renewcommand{\tabularxcolumn}[1]{p{#1}}
\renewcommand{\arraystretch}{1.35}

This section gives the full preprocessing, class-balance sampling, and
optimization settings underlying Section~\ref{sec:training}. Aerial
imagery was cropped to 280~$\times$~280 pixels at 0.1~m ground sampling
distance (Table~\ref{tab:encoders}), resized to 256~$\times$~256, and
normalized with ImageNet channel-wise mean and standard deviation; for
sites with more than one eligible SWISSIMAGE acquisition, training
sampled a leaf-on year at random on each draw, whereas cross-validation
evaluation used the eligible acquisition closest to the site's NFI survey
year. National inference instead used the most recent leaf-on acquisition
available in each area, as described in Section~\ref{sec:inference-data}.
For training and cross-validation, the AlphaEarth patch was taken from
the site's NFI survey year; for national inference, it was taken from the
year of the selected SWISSIMAGE acquisition. The
ALS point cloud was cropped to a 14~m radius, voxelized at 0.1~m, and the
LAS ground-return class (class~2) was removed before converting point
elevations to height above the local terrain using the DEM. AlphaEarth
patches comprised $3\times3$ pixels, giving nine 64-dimensional feature
vectors per site (Table~\ref{tab:encoders}). RGB, the ALS $xy$-coordinates,
and the AlphaEarth patch were augmented with a shared random rotation and
flip per site, so that the three spatially registered inputs remained
aligned.

Rare-species representation was addressed by a distributed rare-site
weighted sampler. Pine, larch, Arolla pine, maple, ash, oak, and chestnut
were designated as rare classes. A site received a sampling weight of 1.5
if any of these classes reached a 5\% basal-area share, regardless of which
class was dominant; all other sites received a weight of 1.0. Sampling was
performed with replacement and affected how often a site was drawn rather
than its contribution to the loss (Section~\ref{sec:model-heads}). In the
single-GPU experiments, each epoch contained as many draws as there were
sites in the corresponding training fold, so individual sites could be
repeated or omitted within an epoch. Whenever a site was sampled again, its
leaf-on SWISSIMAGE acquisition was re-sampled independently.

Apart from the frozen DINOv3 aerial encoder, all components were trained
from random initialization (Table~\ref{tab:encoders}). Optimization
settings are listed in Table~\ref{tab:optimization}.

\begin{table}[pos=htbp]
\centering
\caption{Optimization settings.}
\label{tab:optimization}
\setlength{\tabcolsep}{3pt}
\renewcommand{\arraystretch}{1.15}
\small
\begin{tabularx}{0.6\textwidth}{@{}>{\raggedright\arraybackslash}X
                              >{\raggedright\arraybackslash}m{4cm}}
\toprule
\textbf{Setting} & \textbf{Value} \\
\midrule
Optimizer & AdamW \\
Initial learning rate & $1\times10^{-4}$ \\
Weight decay & 0.05 \\
Adam betas & (0.9, 0.999) \\
Scheduler & Cosine decay \\
Minimum learning rate & $1\times10^{-6}$ \\
Warm-up & None \\
Epochs & 80 \\
Global batch size & 4 \\
Gradient clipping & Max norm 1.0 \\
Mixed precision & Disabled (FP32) \\
EMA & Decay 0.999 \\
Random seed & 42 \\
Loss weight, stand type & 1.0 \\
Loss weights, leaf type and composition (Eq.~\ref{eq:comp-loss}) & $\lambda_{\mathrm{KL}}=0.2$, $\lambda_{L_1}=1.0$ \\
Loss weight, consistency (Eq.~\ref{eq:total-loss}) & 0.1 \\
\bottomrule
\end{tabularx}
\end{table}

\section{Evaluation metrics}
\label{sec:appendix-metrics}

Let $y_{ik}$ and $\hat{y}_{ik}$ denote the observed and predicted
proportions of class $k$ at site $i$ over the $n$ evaluation sites. For each
class, the mean absolute error (MAE), the root mean squared error (RMSE),
and the coefficient of determination ($R^2$) are

\begin{equation}
\mathrm{MAE}_k=\frac{1}{n}\sum_{i=1}^{n}\left|y_{ik}-\hat{y}_{ik}\right|,
\label{eq:mae}
\end{equation}

\begin{equation}
\mathrm{RMSE}_k=\sqrt{\frac{1}{n}\sum_{i=1}^{n}
\left(y_{ik}-\hat{y}_{ik}\right)^2},
\label{eq:rmse}
\end{equation}

\begin{equation}
R^2_k=1-\frac{\sum_{i=1}^{n}\left(y_{ik}-\hat{y}_{ik}\right)^2}
{\sum_{i=1}^{n}\left(y_{ik}-\bar{y}_k\right)^2}.
\label{eq:r2}
\end{equation}

Unless otherwise stated, the metrics reported in
Section~\ref{sec:results} are macro-averages over the $K=11$ mapped
classes, so that every class contributes equally regardless of how abundant
it is,

\begin{equation}
M_{\mathrm{macro}}=\frac{1}{K}\sum_{k=1}^{K}M_k,
\qquad
M\in\left\{\mathrm{MAE},\ \mathrm{RMSE},\ R^2,\ F_1\right\}.
\label{eq:macro}
\end{equation}

For the complementary classification-based evaluation, each observed and
predicted composition vector was converted to its dominant class using

\begin{equation}
c_i=\operatorname*{arg\,max}_k y_{ik},
\qquad
\hat{c}_i=\operatorname*{arg\,max}_k \hat{y}_{ik}.
\label{eq:argmax}
\end{equation}

and the resulting labels were evaluated with overall accuracy and
macro-$F_1$, the latter averaged across the classes represented in the
evaluation dataset.

\begin{figure}[pos=htbp]
\centering
\includegraphics[width=0.667\textwidth]{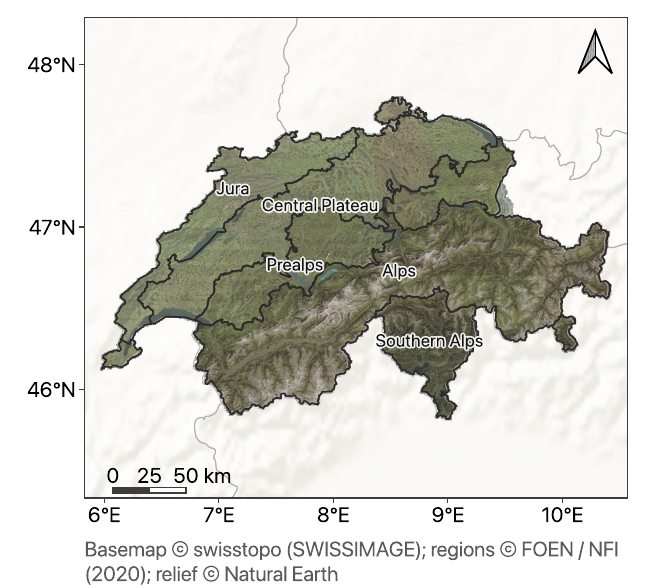}
\caption{The five Swiss NFI production regions used for the comparison
with NFI4 regional statistics (Section~\ref{sec:res-nfi4}).}
\label{fig:production-regions}
\end{figure}

Figure~\ref{fig:production-regions} shows the five NFI production regions
used for the comparison with NFI4 regional statistics.

\section{Alternative design choices}
\label{sec:appendix-design}

Two further groups of experiments examined design choices that were made
within the selected modality set: how satellite information is represented
and how climate context is supplied. Each group replaced a single component
of the reference configuration while leaving everything else unchanged.

The Sentinel-2 alternative used Level-1C imagery
\citep{europeanspaceagency2015sentinel2} processed to bottom-of-atmosphere
(BOA) surface reflectance with FORCE \citep{frantz2019force} on a nominal
10--20~m grid. For each site, a 3~$\times$~3-pixel patch was extracted from
all acquisitions within the site's NFI observation year, each providing the
ten FORCE-derived spectral bands together with its quality mask; cloud,
cloud shadow, snow, saturation, and no-data observations were masked during
training. The resulting seasonal NDVI trajectories of the mapped classes are
shown in Section~\ref{sec:appendix-s2phenology}
(Fig.~\ref{fig:ndvi-phenology}).

Substituting the annual AlphaEarth embeddings with a directly encoded
Sentinel-2 reflectance time series reduced composition $R^2$ from 0.485 to
between 0.458 and 0.480, depending on the temporal window
(Table~\ref{tab:satellite}). Within the raw Sentinel-2 variants the
ordering was interpretable: the full-year series performed best (0.480),
ahead of the single July composite (0.464) and the growing-season series
(0.458), which is consistent with the leaf-emergence and senescence periods
carrying much of the phenological contrast between the mapped classes
(Section~\ref{sec:appendix-s2phenology}) and therefore favoring the
window that spans both. The margin over the best raw variant is only 0.005,
however, so on this evidence the embeddings and a full-year Sentinel-2
series are close to interchangeable for composition; the full-year variant
is in fact ahead on dominant-class $F_1$. Two properties
of the AlphaEarth representation plausibly account for the difference. It
is derived from Sentinel-1, Sentinel-2, and Landsat observations rather
than from Sentinel-2 alone, and it is produced as a gap-free annual
summary, whereas the FORCE series is restricted to the site's inventory
year and is thinned by cloud, snow, and shadow masking.

\begin{table}[pos=htbp]
\caption{Satellite representation. Best in bold, second best underlined. The dagger marks the configuration retained in the reference model.}
\label{tab:satellite}
\footnotesize
\renewcommand{\arraystretch}{1.3}
\begin{tabular*}{\textwidth}{@{\extracolsep{\fill}}llllll@{}}
\toprule
\multirow{2}{*}{\textbf{Variant}} & \multicolumn{2}{c}{\textbf{Leaf level}} & \multicolumn{3}{c}{\textbf{Species level}} \\
\cmidrule(lr){2-3}\cmidrule(l){4-6}
 & \textbf{MAE} & \textbf{$R^2$} & \textbf{MAE} & \textbf{$R^2$} & \textbf{$F_1$} \\
\midrule
AlphaEarth embeddings$^\dagger$ & \textbf{0.093$\pm$0.003} & \textbf{0.803$\pm$0.017} & \textbf{0.048$\pm$0.001} & \textbf{0.485$\pm$0.038} & 0.585$\pm$0.014 \\
Sentinel-2, July composite & 0.098$\pm$0.004 & \underline{0.791$\pm$0.015} & 0.050$\pm$0.001 & 0.464$\pm$0.044 & \underline{0.588$\pm$0.029} \\
Sentinel-2, growing season & 0.099$\pm$0.003 & 0.786$\pm$0.018 & 0.050$\pm$0.001 & 0.458$\pm$0.052 & 0.579$\pm$0.041 \\
Sentinel-2, all year & \underline{0.098$\pm$0.003} & 0.790$\pm$0.011 & \underline{0.050$\pm$0.001} & \underline{0.480$\pm$0.042} & \textbf{0.593$\pm$0.020} \\
\bottomrule
\end{tabular*}
\end{table}
\renewcommand{\arraystretch}{1.35}

Two alternatives to the explicit CHELSA variables were tested
(Table~\ref{tab:climate}). Neither is separated from the CHELSA
configuration by much. The frozen Climplicit embedding was marginally ahead
on every metric except species-level MAE: composition $R^2$ 0.487 against
0.485, leaf-type $R^2$ 0.813 against 0.803, and dominant-class $F_1$ 0.593
against 0.585. The Location-MLP variant, in which the climate branch
received only the site coordinates, was the best of the three on the
leaf-type target (leaf-type $R^2$ 0.815) but the weakest on the composition
target ($R^2$ 0.473), which is the quantity the product is intended to
deliver.

We nevertheless retained the explicit CHELSA variables, and on this evidence
the choice cannot be defended on accuracy. The spread across the three
climate representations is 0.014 in composition $R^2$, and the gap between
CHELSA and Climplicit is 0.002, against the 0.020 cost of removing the
climate branch altogether (Section~\ref{sec:res-modalities}). The two are best regarded as
equivalent. The argument for CHELSA is therefore one of interpretability
and transferability rather than of measured accuracy: its 19 variables are
physically defined and available on the same scale everywhere, so the
fitted model can be applied and audited outside Switzerland, whereas
Climplicit is itself a learned embedding whose behavior outside its
training distribution is not directly inspectable. That a coordinate-only
branch comes within 0.015 of both is separately informative: it indicates
that much of what the climate branch contributes is large-scale spatial
structure, which coordinates encode together with biogeographic region,
management history, and species-pool composition. A coordinate branch,
however, can only reproduce the spatial pattern it was trained on, which is
why it was not carried forward.

\begin{table}[pos=htbp]
\caption{Climate representation. Best in bold, second best underlined. The dagger marks the configuration retained in the reference model.}
\label{tab:climate}
\footnotesize
\renewcommand{\arraystretch}{1.3}
\begin{tabular*}{\textwidth}{@{\extracolsep{\fill}}llllll@{}}
\toprule
\multirow{2}{*}{\textbf{Variant}} & \multicolumn{2}{c}{\textbf{Leaf level}} & \multicolumn{3}{c}{\textbf{Species level}} \\
\cmidrule(lr){2-3}\cmidrule(l){4-6}
 & \textbf{MAE} & \textbf{$R^2$} & \textbf{MAE} & \textbf{$R^2$} & \textbf{$F_1$} \\
\midrule
CHELSA BIOCLIM+$^\dagger$ & 0.093$\pm$0.003 & 0.803$\pm$0.017 & \textbf{0.048$\pm$0.001} & \underline{0.485$\pm$0.038} & 0.585$\pm$0.014 \\
Climplicit embedding & \textbf{0.092$\pm$0.003} & \underline{0.813$\pm$0.015} & \underline{0.049$\pm$0.001} & \textbf{0.487$\pm$0.042} & \textbf{0.593$\pm$0.027} \\
Location MLP & \underline{0.093$\pm$0.003} & \textbf{0.815$\pm$0.014} & 0.049$\pm$0.002 & 0.473$\pm$0.044 & \underline{0.585$\pm$0.023} \\
\bottomrule
\end{tabular*}
\end{table}
\renewcommand{\arraystretch}{1.35}

A pattern recurs across these groups: several alternatives that lose
composition accuracy nonetheless achieve a higher dominant-class $F_1$ than
the reference configuration. The full-year Sentinel-2 variant gives up 0.005
in composition $R^2$ but gains 0.008 in $F_1$, and the same 0.008 gain
appears for the Climplicit climate branch and for the ALS-free modality
combination of Section~\ref{sec:res-modalities}. The dissociation is
therefore consistent in sign but small in size, well below the 0.070
difference in dominant-class $F_1$ between the reference configuration
(0.585) and the strongest non-multimodal baseline (0.515) in
Table~\ref{tab:baselines}. Because $F_1$ is evaluated after the argmax operation, a
configuration can improve it by sharpening the leading component while
distributing the remaining share less accurately. The reference
configuration was identified on the continuous metrics, which are the
quantities the product is intended to deliver.

\section{Independent validation data and protocol}
\label{sec:appendix-reserves}

\subsection*{Validation data}

The Forest Reserves Monitoring network \citep{hobi2020reserves} provides
three linked components: a tree-level inventory table, a species lookup
table, and permanent-plot polygons. The inventory table records, for each
tree and each census, the reserve and plot identifiers, the census year,
the species code, the coordinates, the tree status, the DBH, and the
vertical stratum. The lookup table links species codes to taxon names,
which were then mapped onto the 11 classes of
Table~\ref{tab:species-groups} using the same crosswalk as the training
labels; in particular, \textit{Pinus} spp., \textit{P.\ sylvestris}, and
the two \textit{P.\ mugo} subspecies were assigned to Pine, and
\textit{P.\ cembra} to Arolla pine.

The network was established to study the long-term dynamics of unmanaged
forests and to contrast reserves with managed stands. It is not a
probability sample of Swiss forest, and the assessment reported in
Section~\ref{sec:res-reserves} should accordingly be read as an independent
plot-level evaluation across 35 reserves rather than as a design-unbiased
estimate of national product accuracy. Plots within a reserve are also
spatially clustered, so any uncertainty interval should be obtained by
resampling whole reserves rather than individual plots.

\subsection*{Field composition}

For every plot, only the most recent census was retained, so that repeated
measurements of the same plot are not treated as independent observations.
Trees were then restricted to living individuals of the upper and middle
storeys with DBH $\geq$ 12~cm, which aligns the validation target with the
label definition of Section~\ref{sec:labels}; 28,152 tree records remained.
Applying the 12~cm threshold removed a large number of stems but only about
0.89\% of the total basal area, and it changed the headline metrics only
marginally. The stratum field of the reserve inventory and the NFI
\texttt{SCHICHT} field originate from different inventory systems and are
therefore closely harmonized rather than strictly identical.

Basal area was computed as $BA_i=\pi(d_i/2000)^2$ with $d_i$ the DBH in
millimetres, which is Eq.~\ref{eq:tree-ba} expressed in the units of this
dataset, and the observed proportion of class $k$ in plot $p$ was taken as
the class basal-area sum divided by the plot total. Unlike the NFI, the
reserve inventories are complete enumerations within a fixed polygon and
provide no tree-level expansion factors; all retained trees in a plot
therefore share the same spatial support, and the common plot-area term
cancels when within-plot proportions are formed.

\subsection*{Extraction of the prediction}

Predictions were not read from the pixel containing the plot center.
Instead, each plot polygon was transformed to the coordinate system of the
national product, all valid 30~m forest pixels intersecting the polygon
were identified, and the plot-level prediction was formed as the
intersection-area-weighted mean of their 11-part composition vectors,

\begin{equation}
\hat{y}_{pk}=\frac{\sum_j A_{pj}\,\hat{y}_{jk}}{\sum_j A_{pj}},
\label{eq:plot-weighted}
\end{equation}

where $A_{pj}$ is the area of intersection between plot $p$ and pixel $j$.
The result was renormalized to sum to one, correcting only numerical
departures from closure. Of the 135 plots in the network, 132 had valid
predictions; three were dropped because no valid product pixel overlapped
their polygon. Across the retained plots the mean valid-pixel coverage was
99.92\% and the minimum 96.19\%.

The pixel-level evaluation of Section~\ref{sec:res-reserves} uses the same
intersecting pixels but does not average across them. Each intersecting
pixel was instead treated as an independent observation if the plot
polygon covered at least 80\% of its area and at least 10 retained trees
fell within it; eligible pixels received the prediction of their own
raster cell directly, with no spatial averaging. Because a retained
tree's coordinates can fall just outside the plot polygon through small
GPS or digitization error, a tree was still counted toward a pixel if it
lay within 5~m of the polygon boundary; this tolerance retained 98.85\%
of trees (99.22\% of basal area), against 94.41\% (95.26\% of basal area)
with no tolerance. Under this specification, 414 pixels from 111 plots in
34 reserves were retained, with a median of 29 retained trees per
eligible pixel.

\subsection*{Threshold sensitivity}

\begin{figure}[pos=htbp]
\centering
\includegraphics[width=0.8\textwidth]{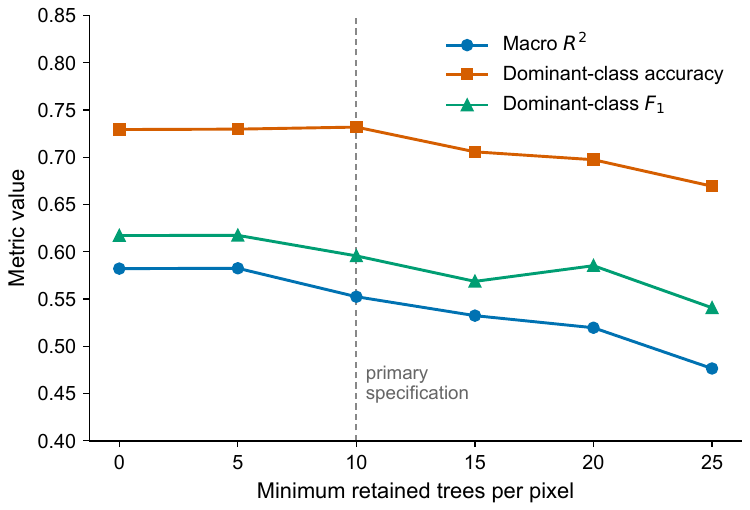}
\caption{Sensitivity of the pixel-level validation to the
minimum-tree-count threshold, with pixel coverage fixed at 80\%. The
dashed line marks the primary specification (10 trees) used in
Section~\ref{sec:res-reserves}.}
\label{fig:pixel-threshold}
\end{figure}

The 80\% coverage and 10-tree thresholds were set by a sensitivity
analysis rather than chosen post hoc to favor a particular result.
Holding coverage at 80\% and varying only the minimum tree count
(Fig.~\ref{fig:pixel-threshold}), performance was stable from no explicit
minimum up to 10 trees: macro $R^2$ fell by only 0.03 (from 0.582 to
0.552) and dominant-class accuracy was essentially unchanged. Stricter
thresholds were markedly more costly: at 25 trees only 248 pixels from 85
plots and 28 reserves remained (56\% of the 10-tree sample), and macro
$R^2$ fell to 0.477. Across coverage thresholds of 80--95\% with the
10-tree minimum held fixed, macro $R^2$ stayed within a narrow band of
approximately 0.53--0.55. The primary specification (80\% coverage, 10
trees) was retained as a compromise between the stability of the
pixel-level reference composition and the size of the validation sample,
not as an empirically optimized choice.

\subsection*{Metrics}

MAE, RMSE, and $R^2$ follow Eqs.~\ref{eq:mae}--\ref{eq:r2}, with the plot
as the unit of observation for the plot-level evaluation and the pixel as
the unit of observation for the pixel-level evaluation, and are
aggregated as in Eq.~\ref{eq:macro}.
Because every class is observed at the same 132 plots, the macro-averaged
and pooled MAE coincide; the macro-averaged and pooled RMSE do not, and the
pooled value of 0.098 is retained only as a diagnostic. The per-class
results are given in Table~\ref{tab:reserve-perclass}.

Macro-averaged MAE and RMSE retained all 11 mapped classes because neither
metric depends on reference variance. The treatment of chestnut matters
more for $R^2$. Chestnut was sufficiently represented in the spatial
cross-validation, and including it changed the macro-$R^2$ only slightly,
from 0.483 to 0.485 (Table~\ref{tab:r2-chestnut-sensitivity}). In the
independent data, however, chestnut occurred with a non-zero share at only
one of the 132 plots and at none of the 414 eligible pixels.

At plot support, chestnut's single non-zero observation (0.00397) yielded
$\mathrm{SST}=1.566\times10^{-5}$ and
$\mathrm{SSE}=7.436\times10^{-4}$, giving a diagnostic $R^2$ of
approximately $-46.5$. Including it would reduce the plot-level
macro-$R^2$ from 0.733 to $-3.560$. At pixel support, all reference
chestnut shares were zero, so $R^2$ was undefined. We therefore used the
same ten evaluable classes when comparing $R^2$ across the three supports.

\begin{table}[pos=htbp]
\caption{Sensitivity of macro-averaged $R^2$ to the inclusion of chestnut
across evaluation supports. Cross-validation values are means of the five
fold-level macro-$R^2$ values. \textsuperscript{a} At plot support, the value
including chestnut is reported only as a diagnostic because chestnut has a
single non-zero reference observation. \textsuperscript{b}At pixel support,
chestnut has zero reference variance and $R^2$ is therefore not evaluable
(n.e.). The second row uses the same ten evaluable classes at all three
supports.}
\label{tab:r2-chestnut-sensitivity}
\centering
\small
\renewcommand{\arraystretch}{1.25}
\setlength{\tabcolsep}{5pt}
\begin{tabular}{@{}lccc@{}}
\toprule
\textbf{Class set} & \textbf{Spatial 5-fold CV} & \textbf{Independent, plot} & \textbf{Independent, pixel} \\
 & NFI ($n=3{,}992$) & Forest reserve ($n=132$) & Forest reserve ($n=414$) \\
\midrule
All 11 classes (including chestnut) & 0.485 & $-3.560$\textsuperscript{a} & n.e.\textsuperscript{b} \\
Ten evaluable classes (excluding chestnut) & 0.483 & 0.733 & 0.552 \\
\bottomrule
\end{tabular}
\end{table}
\renewcommand{\arraystretch}{1.35}

\begin{table}[pos=htbp]
\caption{Per-class performance at the plot-level (132 forest reserve
plots) and pixel-level (414 forest-reserve pixels) independent
validations. $n_+$ is the number of plots or pixels in which the class
was observed with a non-zero share. Chestnut has a single non-zero
plot-level observation and no positive pixel-level observation, so its
$R^2$ is not evaluable (n.e.) at either support and it is excluded from
the macro-$R^2$; the macro-MAE and macro-RMSE retain all 11 classes.
Dominant-class support at each observation unit is given in the row labels
of Fig.~\ref{fig:reserve-confusion}.}
\label{tab:reserve-perclass}
\footnotesize
\renewcommand{\arraystretch}{1.3}
\begin{tabular*}{\textwidth}{@{\extracolsep{\fill}}lcccccccc@{}}
\toprule
\multirow{2}{*}{\textbf{Class}} & \multicolumn{2}{c}{\textbf{MAE}} & \multicolumn{2}{c}{\textbf{RMSE}} & \multicolumn{2}{c}{\textbf{$R^2$}} & \multicolumn{2}{c}{\textbf{$n_+$}} \\
\cmidrule(lr){2-3}\cmidrule(lr){4-5}\cmidrule(lr){6-7}\cmidrule(l){8-9}
 & \textbf{Plot} & \textbf{Pixel} & \textbf{Plot} & \textbf{Pixel} & \textbf{Plot} & \textbf{Pixel} & \textbf{Plot} & \textbf{Pixel} \\
\midrule
Spruce      & 0.085 & 0.115 & 0.172 & 0.237 & 0.668 & 0.677 & 89 & 264 \\
Fir         & 0.034 & 0.038 & 0.071 & 0.092 & 0.829 & 0.643 & 48 & 100 \\
Pine        & 0.044 & 0.063 & 0.135 & 0.194 & 0.718 & 0.421 & 50 & 102 \\
Larch       & 0.020 & 0.029 & 0.058 & 0.085 & 0.869 & 0.728 & 23 & 61 \\
Arolla pine & 0.031 & 0.039 & 0.118 & 0.138 & 0.700 & 0.610 & 18 & 56 \\
Beech       & 0.058 & 0.063 & 0.099 & 0.125 & 0.880 & 0.811 & 74 & 175 \\
Maple       & 0.024 & 0.025 & 0.042 & 0.062 & 0.633 & 0.021 & 74 & 124 \\
Ash         & 0.026 & 0.018 & 0.066 & 0.059 & 0.766 & 0.647 & 53 & 73 \\
Oak         & 0.043 & 0.034 & 0.098 & 0.108 & 0.830 & 0.649 & 46 & 76 \\
Chestnut    & 0.001 & 0.000 & 0.002 & 0.002 & n.e.  & n.e.  & 1  & 0 \\
Other       & 0.048 & 0.036 & 0.104 & 0.100 & 0.439 & 0.317 & 81 & 115 \\
\midrule
Macro       & 0.038 & 0.042 & 0.088 & 0.109 & 0.733 & 0.552 & --- & --- \\
\bottomrule
\end{tabular*}
\end{table}
\renewcommand{\arraystretch}{1.35}

\subsection*{Comparison with cross-validation}

The cross-validation metrics reported in
Table~\ref{tab:support-comparison} follow the same per-fold,
mean-across-folds procedure used for Tables~\ref{tab:baselines}
and~\ref{tab:modalities}, applied to the reference configuration. Within each fold, MAE and RMSE were
macro-averaged over all 11 mapped classes, dominant-class accuracy was
calculated over all observations, and macro-$F_1$ retained all 11
represented dominant classes. For the support-level $R^2$ comparison in
Section~\ref{sec:res-reserves}, chestnut was excluded so that the same ten
evaluable classes were used at all three supports
(Table~\ref{tab:r2-chestnut-sensitivity}). Each reported cross-validation
value is the mean of the five resulting fold-level values.

Table~\ref{tab:cv-pixel-species} compares the per-class result of this
cross-validation with the pixel-level independent validation. Agreement
is not uniform across classes. Beech and larch have low MAE at both
supports, and larch also has similar $R^2$ values, whereas beech's $R^2$
increases from 0.673 in cross-validation to 0.811 in the pixel-level
validation. Pine shows the clearest consistent deterioration: MAE, RMSE,
and dominant-class recovery (Section~\ref{sec:res-reserves}) are all worse
at the pixel level.
Several other classes show large $R^2$ shifts driven by the reference
composition rather than by prediction error: maple's MAE and RMSE are
both smaller at the pixel level (0.025 and 0.062, against 0.051 and
0.124 in the cross-validation) yet its $R^2$ falls from 0.258 to 0.021,
because the standard deviation of the pixel-level reference share is
much smaller than that of the cross-validation sites. This is the same
mechanism documented for chestnut above: $R^2$ is one minus a residual
sum of squares divided by the observed variance (Eq.~\ref{eq:r2}), so a
smaller observed variance can lower $R^2$ even when the absolute error
improves.

\begin{table}[pos=htbp]
\caption{Per-class MAE and $R^2$ in the spatial five-fold
cross-validation and the pixel-level independent validation.
Cross-validation values are the mean over the five folds, restricted to
the reference configuration. Chestnut is omitted from this side-by-side
table because its pixel-level $R^2$ is not evaluable.}
\label{tab:cv-pixel-species}
\footnotesize
\renewcommand{\arraystretch}{1.3}
\begin{tabular*}{\textwidth}{@{\extracolsep{\fill}}lllll@{}}
\toprule
\multirow{2}{*}{\textbf{Class}} & \multicolumn{2}{c}{\textbf{MAE}} & \multicolumn{2}{c}{\textbf{$R^2$}} \\
\cmidrule(lr){2-3}\cmidrule(l){4-5}
 & \textbf{CV} & \textbf{Pixel-level} & \textbf{CV} & \textbf{Pixel-level} \\
\midrule
Spruce      & 0.111 & 0.115 & 0.746 & 0.677 \\
Fir         & 0.061 & 0.038 & 0.668 & 0.643 \\
Pine        & 0.024 & 0.063 & 0.599 & 0.421 \\
Larch       & 0.039 & 0.029 & 0.742 & 0.728 \\
Arolla pine & 0.013 & 0.039 & 0.339 & 0.610 \\
Beech       & 0.084 & 0.063 & 0.673 & 0.811 \\
Maple       & 0.051 & 0.025 & 0.258 & 0.021 \\
Ash         & 0.037 & 0.018 & 0.242 & 0.647 \\
Oak         & 0.028 & 0.034 & 0.236 & 0.649 \\
Other       & 0.073 & 0.036 & 0.331 & 0.317 \\
\bottomrule
\end{tabular*}
\end{table}
\renewcommand{\arraystretch}{1.35}

\end{document}